\documentclass{article}
\usepackage{PRIMEarxiv}

\usepackage{times}  
\usepackage{helvet}  
\usepackage{courier}  
\usepackage[hyphens]{url}  
\usepackage{graphicx} 
\usepackage[sort&compress, numbers]{natbib}
\usepackage{caption} 

\usepackage{fancyhdr}       
\usepackage{enumitem}

\usepackage{amsmath}
\usepackage{amssymb}
\usepackage[show]{chato-notes}
\usepackage{enumitem} 

\usepackage{mathtools}
\usepackage{amsthm}

\usepackage[capitalize,noabbrev]{cleveref}

\theoremstyle{plain}
\newtheorem{theorem}{Theorem}[section]

\newtheorem{lemma}[theorem]{Lemma}

\theoremstyle{definition}
\newtheorem{definition}[theorem]{Definition}

\theoremstyle{remark}

\usepackage{textcomp}
\usepackage{xcolor}
\usepackage{booktabs} 
\usepackage[utf8]{inputenc}
\usepackage{multirow}

\usepackage{xspace}
\usepackage{float}

\usepackage{algorithmic}

\usepackage{amsthm}
\usepackage{amsfonts}

\usepackage{subcaption}

\usepackage{bm}
\usepackage{mathtools}
\usepackage{stmaryrd}
\usepackage{stmaryrd}
\usepackage[disable]{todonotes}

\newtheorem{problem}{Problem}
\newtheorem{property}{Property}

\usepackage{xcolor} 

\newcommand{\squishlist}{
 \begin{list}{$\bullet$}
  {  \setlength{\itemsep}{0pt}
     \setlength{\parsep}{3pt}
     \setlength{\topsep}{3pt}
     \setlength{\partopsep}{0pt}
     \setlength{\leftmargin}{2em}
     \setlength{\labelwidth}{1.5em}
     \setlength{\labelsep}{0.5em}
} }
\newcommand{\squishlisttight}{
 \begin{list}{$\bullet$}
  { \setlength{\itemsep}{0pt}
    \setlength{\parsep}{0pt}
    \setlength{\topsep}{0pt}
    \setlength{\partopsep}{0pt}
    \setlength{\leftmargin}{2em}
    \setlength{\labelwidth}{1.5em}
    \setlength{\labelsep}{0.5em}
} }

\newcommand{\squishdesc}{
 \begin{list}{}
  {  \setlength{\itemsep}{0pt}
     \setlength{\parsep}{3pt}
     \setlength{\topsep}{3pt}
     \setlength{\partopsep}{0pt}
     \setlength{\leftmargin}{1em}
     \setlength{\labelwidth}{1.5em}
     \setlength{\labelsep}{0.5em}
} }

\newcommand{\squishend}{
  \end{list}
}

\newcommand{\spara}[1]{\smallskip\noindent{\bf #1}}

\newcommand{\reals}{\ensuremath{\mathbb R}\xspace}

\newcommand{\ssq}{\ensuremath{\lambda_{\max}}}
\newcommand{\parv}{\ensuremath{\mathbf{v}(\mu)}}

\newcommand{\lagrangian}{\ensuremath{\mathcal{H}}}

\newcommand{\matrixX}{\ensuremath{\mathbf{X}}\xspace}

\newcommand{\matrixI}{\ensuremath{\mathbf{I}}\xspace}
\newcommand{\matrixM}{\ensuremath{\mathbf{M}}\xspace}
\newcommand{\Ait}{\ensuremath{\mathbf{A}_i^{\top}\mathbf{A}_i}\xspace}
\newcommand{\bfA}{\ensuremath{\mathbf{A}}\xspace}
\newcommand{\bfB}{\ensuremath{\mathbf{B}}\xspace}
\newcommand{\bfV}{\ensuremath{\mathbf{V}}\xspace}

\newcommand{\bfU}{\ensuremath{\mathbf{U}}\xspace}

\newcommand{\bfP}{\ensuremath{\mathbf{P}}\xspace}
\newcommand{\matrixC}{\ensuremath{\mathbf{C}}\xspace}
\newcommand{\bfs}{\ensuremath{\mathbf{s}}\xspace}
\newcommand{\bfone}{\ensuremath{\mathbf{1}}\xspace}
\newcommand{\bfy}{\ensuremath{\mathbf{y}}\xspace}
\newcommand{\bfx}{\ensuremath{\mathbf{x}}\xspace}

\newcommand{\VV}{\ensuremath\mathcal V}
\newcommand{\groups}{\ensuremath\mathcal G}

\newcommand{\Vc}{\ensuremath{\bfV^{\perp}}}

\newcommand{\bfmu}{\ensuremath{\boldsymbol{\mu}}\xspace}
\newcommand{\bfv}{\ensuremath{\mathbf{v}}\xspace}
\newcommand{\bfw}{\ensuremath{\mathbf{w}}\xspace}

\newcommand{\svd}{{\sc SVD}\xspace}

\newcommand{\pca}{{\sc PCA}\xspace}

\newcommand{\loss}{{\ensuremath {\mathcal{L}} }\xspace}

\newcommand{\communities}{{\sf\small communities}\xspace}
\newcommand{\communitiesfour}{{\sf\small communities-4}\xspace}
\newcommand{\compas}{{\sf\small compas}\xspace}
\newcommand{\compasthree}{{\sf\small compas-3}\xspace}
\newcommand{\adult}{{\sf\small adult}\xspace}
\newcommand{\german}{{\sf\small german}\xspace}
\newcommand{\recidivism}{{\sf\small recidivism}\xspace}
\newcommand{\student}{{\sf\small student}\xspace}

\newcommand{\heart}{{\sf\small heart}\xspace}
\newcommand{\credit}{{\sf\small credit}\xspace}

\newcommand{\gaussianthree}{{\sf\small gaussian-3}\xspace}

\newcommand{\A}{\ensuremath{{A}}\xspace}
\newcommand{\B}{\ensuremath{{B}}\xspace}

\newcommand{\fairPCA}{{\sc
Fair \pca}\xspace}

\newcommand{\fairsvd}{{\sc Fair PCs}\xspace}
\newcommand{\faireig}{{\sc Fair-PC}\xspace}
\newcommand{\dual}{{\sc FAIR-PC-Dual}\xspace}

\newcommand{\fairpca}{{\sc FAIR-PCA}\xspace}
\newcommand{\fairpcaalgo}{{\sc FAIR-PCA-SDP}\xspace}
\newcommand{\bicriteria}{{\sc Bicriteria}\xspace}

\newcommand{\sdp}{{\sc SDP}\xspace}

\usepackage{algorithm}
\usepackage{algorithmic}

\usepackage{newfloat}
\usepackage{listings}
\DeclareCaptionStyle{ruled}{labelfont=normalfont,labelsep=colon,strut=off} 
\floatstyle{ruled}
\newfloat{listing}{tb}{lst}{}
\floatname{listing}{Listing}
\title{Fair PCA, One Component at a Time}

\author{
  Antonis Matakos \\
  Aalto University \\
  Espoo, Finland \\
  \texttt{antonis.matakos@aalto.fi} \\
   \And
  Martino Ciaperoni \\
  Aalto University \\
  Espoo, Finland \\
  \texttt{martino.ciaperoni@aalto.fi} \\
  \And
  Heikki Mannila \\
  Aalto University \\
  Espoo, Finland \\
  \texttt{heikki.mannila@aalto.fi}
}

\begin{document}

\maketitle

\begin{abstract}
 The Min-Max \fairpca problem seeks a low-rank representation of multi\-group data such that the the approximation error is as balanced as possible across groups. 
 Existing approaches to this problem return a rank-$d$ fair subspace, but lack the fundamental \emph{containment} property of standard \pca: each rank-$d$ \pca subspace should contain all lower-rank \pca subspaces.
 To fill this gap, we define fair principal components as directions that minimize the maximum group-wise reconstruction error, subject to orthogonality with previously selected components, and we introduce an iterative method to compute them.
 This approach preserves the containment property of standard \pca, and reduces to standard \pca for data with a single group.
 We analyze the theoretical properties of our method and show empirically that it outperforms existing approaches to Min-Max \fairpca.
\end{abstract}

\section{Introduction}

Principal Component Analysis (\pca) provides dimensionally reduced representations of data by expressing the data matrix as a linear combination of a small number of factors. \pca is a foundational technique in machine learning and data science, due to the benefits it offers in terms of scalability, interpretability, and its strong mathematical underpinnings.

\pca identifies a sequence of orthonormal vectors, called principal components, that identify directions of maximum variance in the data. 
In particular, the $i$-th vector captures the direction that best reconstructs the data while remaining orthogonal to the first $i-1$ components. 

Dimensionality reduction is achieved by projecting the data onto the subspace spanned by 
a subset of principal components. 
Specifically, a rank-$d$ representation of the data is obtained by projection onto the first $d$ principal components. 
Since each additional component builds upon the previous ones, these subspaces are nested: a rank-$d$ PCA solution contains all lower-rank solutions. In this work, we 
refer to this property as the \emph{containment} property.


In many applications, the rows of a data matrix are grouped based on attributes such as gender or race.  
In such settings, standard PCA may disproportionately represent dominant groups, leading to biased or unfair outcomes.
To address this, prior work ~\cite{samadi2018price,tantipongpipat2019multi,song2024socially} has considered \emph{fair} \pca, which seeks a common subspace that minimizes the worst-case reconstruction error across all groups. 

While these methods ensure fairness at a fixed dimensionality, they lack the containment property: the fair subspace of rank-$d$ does not contain the fair subspaces of lower ranks. As a result, these approaches are not only less flexible and harder to scale—requiring a separate optimization for each rank—but also deviate from the spirit of standard \pca, which constructs a sequence of principal components that can be incrementally extended or truncated.

We propose a new formulation of Fair \pca that, like standard \pca, produces a sequence of components satisfying the containment property. Our method incrementally constructs an orthonormal basis, where each vector is chosen to minimize the maximum reconstruction error across all groups in the data, while remaining orthogonal to previously selected components. We refer to these as \emph{fair principal components}. 


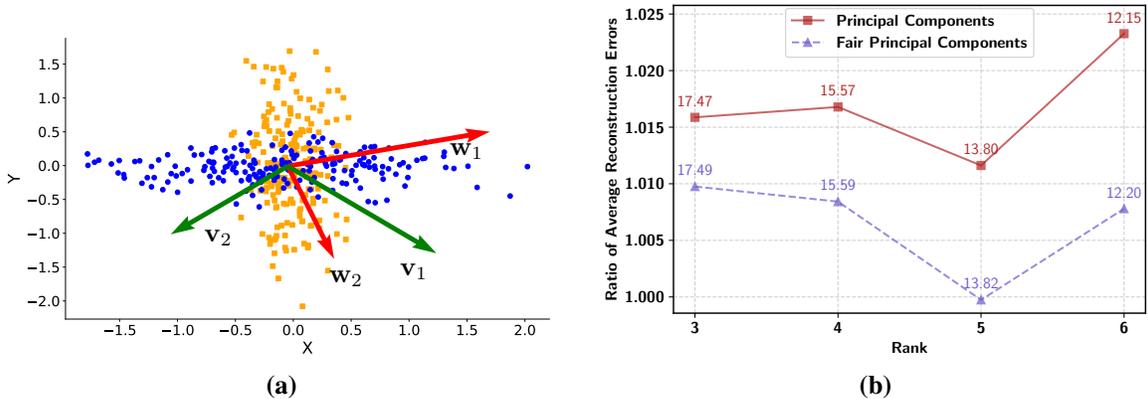
\begin{figure*}[t]
    \centering
    \begin{tabular}{cc}    \includegraphics[width=0.45\textwidth]{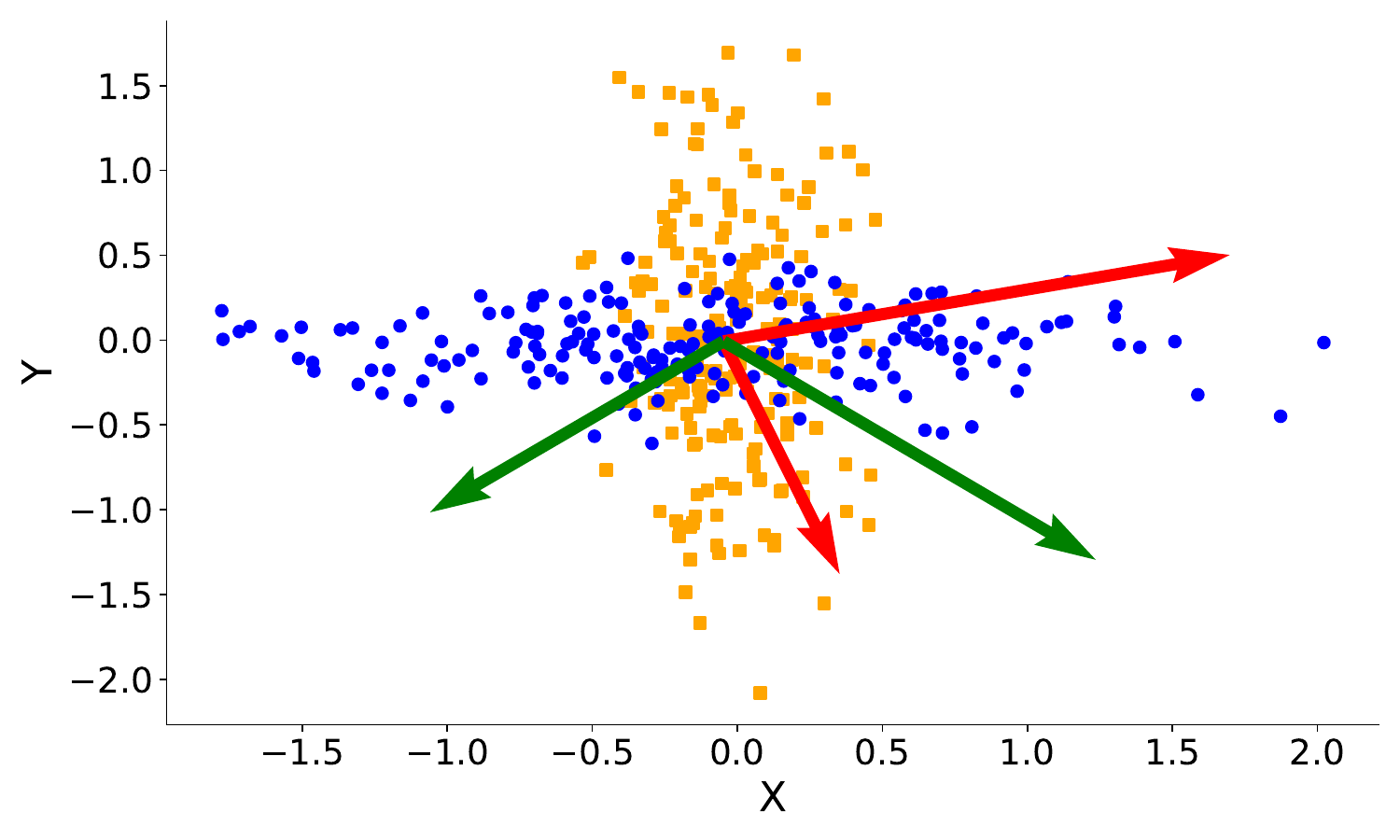}
    \begin{tikzpicture}[overlay]
        \node at (-2,1.25) {$\bfv_1$}; 
        \node at (-4.6,1.7) {$\bfv_2$}; 
        \node at (-1.3,2.85) {$\bfw_1$}; 
        \node at (-2.9,1.15) {$\bfw_2$}; 
    \end{tikzpicture}
    & \includegraphics[width=0.45\textwidth]{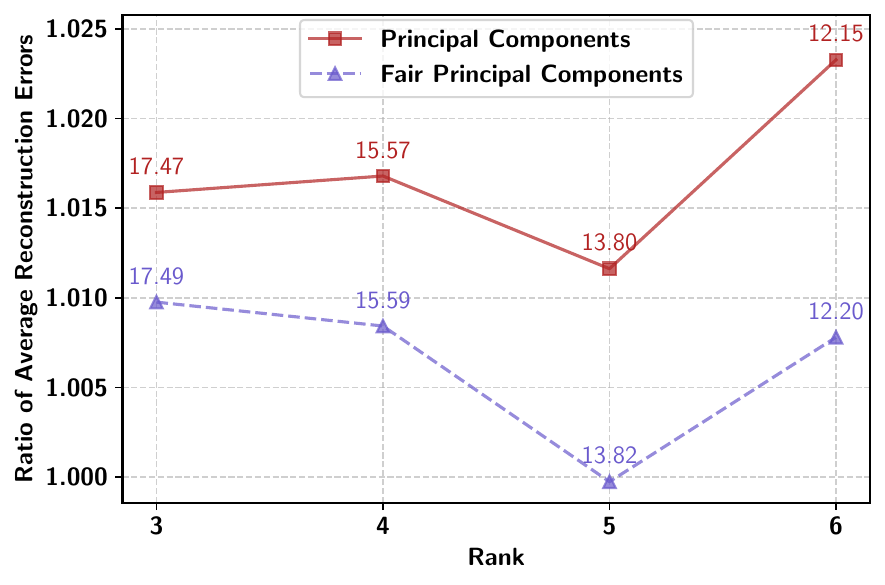}
    \\ 
    \textbf{(a)} & \textbf{(b)} \\ 
    \end{tabular}
    \caption{Left (\textbf{a}): synthetic data partitioned in two groups, as indicated by the color of the points. $\{\bfw_1,\bfw_2\}$ are the standard principal components while $\{\bfv_1,\bfv_2\}$ are the fair principal components given by our method.
    Right (\textbf{b}): real-world \compas dataset partitioned in two groups, females and males. The $y$-axis indicates the ratio of the average group-wise reconstruction error incurred by standard principal components and the fair principal components.
    The $x$-axis indicates the number of components. 
    We also report 
    the average reconstruction error across both groups (males and females).}
    %
\label{fig:two_figures}
\end{figure*}


Figure~\ref{fig:two_figures} illustrates this concept. In panel (a), we show standard principal components (in red) and fair principal components (in green) on synthetic data. Standard PCA favors the majority group, while our method provides a more balanced representation. Panel (b) shows results on the real-world \compas dataset~\cite{dua2019uci}, partitioned by sex. Projecting onto fair principal components yields more balanced reconstruction errors across groups while maintaining similar overall error to standard PCA.


The containment property can be very useful in practical applications (e.g., for feature selection) since, 
once the full-rank basis is computed, lower-dimensional fair subspaces can be obtained simply by discarding components—just as in standard PCA.
Further, a key advantage of our approach is its scalability and efficiency. By incrementally computing one fair principal component at a time, our method decomposes the rank-$d$ problem into $d$ simpler rank-$1$ problems that can be solved efficiently. 

While our method is scalable and modular, computing each fair principal component remains a nontrivial problem. To address this, we develop a primal-dual analysis that reveals a remarkable insight: each fair principal component can be characterized as the leading eigenvector of a carefully chosen convex combination of the group-wise covariance matrices. This mirrors standard PCA, where components correspond to the leading eigenvectors of the overall covariance. Leveraging this connection, we prove that our method is provably optimal in the two-group setting and empirically demonstrate that it achieves near-optimal performance across a range of multi-group scenarios.
Finally, through extensive experiments, we demonstrate that 
our method generally strikes a more desirable balance between efficiency and solution quality than existing methods. 

The contributions of this work can be summarized as follows. 
\begin{itemize}[noitemsep,topsep=0pt, leftmargin=10pt]
    \item We formalize the problem of identifying fair principal components.
     \item We design an iterative procedure which selects the fair principal components according to the min-max criterion, and then projects the data onto the orthogonal complement of the previously chosen fair principal components. The selection of the fair principal component (\faireig) at each iteration represents the main algorithmic challenge of this work.
    \item We present a novel primal-dual analysis for the formulated problem, and we theoretically study the proposed algorithms, focusing mostly on the two-groups case, which exhibits interesting properties. 
    \item We describe extensive experiments on real-world datasets to demonstrate the benefits of our method over previous work.
\end{itemize}

The rest of this paper is organized as follows. Section \ref{sec:related_work} reviews related work. Section \ref{sec:preliminaries} introduces notation and background. Section \ref{sec:overview} describes our framework, and Section \ref{sec:fair_eig} formalizes the \faireig problem. Section \ref{sec:algos} proposes algorithms to solve it, and Section~\ref{sec:twogroup_analysis} gives a theoretical analysis and algorithm for the two-group case. Section \ref{sec:experiments} contains our experimental evaluation. Section~\ref{sec:limitations} discusses some limitations of our work. We conclude with Section~\ref{sec:conclusion}.

\section{Related Work}\label{sec:related_work}

Our work intersects with prior research along several dimensions. 

\spara{Min-Max Fairness.}
A common approach to algorithmic fairness is to ensure equitable performance across different groups defined by sensitive attributes (e.g., gender or race). One widely used framework is Min-Max fairness, which optimizes the worst-case group outcome \cite{abernethy2020active,kandasamy2020online,bistritz2020fair,martinez2020minimax,wang2025stable}. This principle is often associated with the \emph{egalitarian} or \emph{Rawlsian} rule~\cite{sen2017collective}.

\spara{Min-Max Fair PCA.}
Recent work has explored \pca through the lens of Min-Max fairness, with the goal of learning a shared subspace that balances variance across different groups.
This problem, known as \fairPCA~\cite{samadi2018price,tantipongpipat2019multi,shen2025hidden,meng2024efficient}  or "socially fair low-rank approximation"~\cite{song2024socially}, seeks a low-dimensional representation that minimizes the maximum group-wise reconstruction error. Related approaches include the signal processing-based formulation in~\cite{zalcberg2021fair}, and a minorization-maximization strategy proposed by~\citet{babu2025fair}.

\spara{Alternative Fair PCA Formulations.}
Beyond Min-Max objectives, other formulations of fair PCA draw on fairness criteria from supervised learning. For example, several works incorporate notions like \emph{demographic parity} into unsupervised settings~\cite{olfat2019convex,kleindessner2023efficient}. \citet{lee2022fast} define fairness through maximum mean discrepancy between the reduced distributions of different classes. Others, including \citet{pelegrina2023novel} and \citet{kamani2022efficient}, frame fair PCA as a bi-objective optimization that balances accuracy and fairness. 
  
\spara{Fair Dimensionality Reduction.}
Efforts to promote fairness extend beyond PCA to broader dimensionality-reduction techniques. For instance, \citet{matakos2023fair} and \citet{song2024socially} study fair versions of the column subset selection problem, while \citet{louizosSLWZ15} introduce a fair variant of the variational autoencoder.

\section{Preliminaries}
\label{sec:preliminaries}

\spara{Notation.} We denote matrices and vectors by  bold uppercase and lowercase letters, respectively. 
Given a matrix $\bfA \in \reals^{a \times n}$ and a unit vector $\bfv \in \reals^{n}$, the projection onto the orthogonal complement of $\bfv$ is obtained as $\bfA-\bfA\bfv\bfv^{\top}$.
We denote the leading eigenvalue of a symmetric matrix $\bfA$ by $\lambda_{\max}(\bfA)$.  
The Frobenius norm of $\bfA \in \reals^{m \times n}$ is: $\|\bfA\|_F = \sqrt{\sum_{i=1}^m \sum_{j=1}^n |a_{ij}|^2}$, where $a_{ij}$ is the $(i,j)$-th entry of $\bfA$. We assume that \emph{group-wise} matrices $\bfA_1, \ldots, \bfA_k$ are centered independently.


Orthogonal projections satisfy the following useful property (for proof see Appendix~\ref{app:proofs}). 

\begin{property}[Orthogonal projection]
\label{property:projection}
Let $\bfA \in \reals^{m \times n}$ and $\bfV \in \reals^{n \times d}$ have orthonormal columns $\bfv_1, \ldots, \bfv_d$. Then, $\|\bfA \bfV \bfV^\top\|_F^2 = \sum_{i=1}^d \|\bfA \bfv_i \bfv_i^\top\|_F^2 =\sum_{i=1}^d \bfv_i^{\top} \bfA^{\top}\bfA \bfv_i.$
\end{property}
 
A key property of \pca is the \emph{containment} property, which stems from the Eckart–Young–Mirsky theorem~\cite{eckart1936approximation}:  
Let $\matrixM\in \reals^{m\times n}$ with singular value decomposition $\matrixM=\bfU\Sigma\bfV^\top$. Projecting onto the first $d$ singular vectors in $\bfV$, for any $d$, gives the best rank-$d$ approximation to $\matrixM$ under either the Frobenius norm or spectral norm. 

Throughout this work, we assume standard familiarity with \pca; see, e.g., \cite{van2009dimensionality,hotelling1933analysis} for introductions.

\section{Overview of the Method}
\label{sec:overview}

In this section we provide an overview of our approach to fair \pca. 
The core idea is the notion of a \emph{fair principal component}—a direction that defines a rank-1 projection of the data while accounting for fairness across all groups.
Once such a component is computed, we iteratively remove its influence from the data to obtain a sequence of fair components that satisfies the containment property.

We begin by defining the fair principal component problem, called \faireig. Then, we describe an algorithm that computes a sequence of components by solving a series of \faireig problems.

\spara{Fair Principal Component.}  
Let $\matrixM \in \reals^{m \times n}$ be a data matrix with rows partitioned into $k$ groups $\groups = \{\bfA_1, \ldots, \bfA_k\}$. Our goal is to find a direction $\bfv$ that captures as much variance as possible for all groups. 
To this end, we minimize a loss function that measures the worst-case deviation from maximum group-specific variance. Specifically, for a unit vector $\bfv$, the loss is defined as:
\begin{equation}
\label{eq:rank_1}
   \loss(\matrixM, \bfv) = \max_{\bfA_i \in \groups} \left\{ \ssq(\Ait) - \bfv^\top \Ait \bfv \right\},
\end{equation}
where $\ssq(\Ait)$ denotes the maximum variance captured by any rank-1 projection of $\bfA_i$, and $\bfv^\top \Ait \bfv$ is the variance captured by projecting $\bfA_i$ onto $\bfv$. Minimizing this loss ensures that no group is significantly underrepresented relative to its own best-case reconstruction. This objective is closely related to the marginal loss of~\citet{samadi2018price}; see Appendix~\ref{app:loss_functions} for further discussion.

\spara{Computing a Sequence of Fair Principal Components.}  
To construct a sequence of fair principal components, we start by minimizing Equation~\eqref{eq:rank_1} to obtain the first component $\bfv_1$. We then project all group matrices in $\groups$ onto the orthogonal complement of $\bfv_1$, denoted $\{\bfv_1\}^\perp$, and repeat the process to obtain $\bfv_2$. After $d$ iterations, this yields an orthonormal basis $\bfV = \{\bfv_1, \ldots, \bfv_d\}$ of fair components.

This procedure is summarized in Algorithm~\ref{alg:fairsvd}. Step~\ref{step:faireig} calls a subroutine to solve the \faireig problem, while Step~\ref{step:orthcompl} performs the orthogonal projection of group matrices.

\spara{Quality of the Solution.}  
Since each component is orthogonal to the previous ones, Property~\ref{property:projection} and a straightforward inductive argument imply that the total loss for the $d$-dimensional solution is simply the sum of the individual rank-1 losses: $\sum_{i=1}^d \loss(\matrixM, \bfv_i)$.
We refer to this quantity as the \emph{incremental error}, which serves as our primary measure of reconstruction quality. The effectiveness of the overall method thus depends on the quality of the solutions to each rank-1 \faireig problem.

In the following sections, we show that for two groups, \faireig can be solved exactly and efficiently. For more than two groups, we introduce an approximate algorithm that performs well in practice.

\spara{Computational Complexity.}  
The overall time complexity of the method is $\mathcal{O}(d\ell)$, where $\mathcal{O}(\ell)$ is the cost of solving a single \faireig problem. We discuss this subroutine in detail in the next section.

\begin{algorithm}[t!]
\footnotesize
\caption{Fair Orthonormalization}
    \label{alg:fairsvd}
     \begin{algorithmic}[1]
    \STATE  \textbf{Input:} Matrices $\{\bfA_1, \ldots, \bfA_k\}$, rank $d$. 
     \STATE Initialize $r\gets 1$, $V \gets \emptyset$
        \WHILE{$r<=d$}
        \STATE \label{step:faireig} $\bfv_r\gets$ \faireig($\bfA_1,\ldots,\bfA_k$)
        \STATE \label{step:orthcompl} $\bfA_i \gets \bfA_i - \bfA_i\bfv_r\bfv_r^\top$  
        \STATE $\bfV \gets \bfV \cup \bfv_r$
        \STATE $r \gets r+1$
        \ENDWHILE
	\end{algorithmic}
	{\bf return} $\bfV$
\end{algorithm}

\section{The \faireig Problem}
\label{sec:fair_eig}

As outlined in Section~\ref{sec:overview}, the core algorithmic challenge in our method is solving the \faireig problem. In this section, we formalize the problem, analyze its structure, and derive a dual formulation that enables practical optimization. These insights will guide the algorithms introduced in the next section.

\begin{problem}[\faireig]
\label{prob:faireig_optimization}
Given a data matrix $\matrixM \in \reals^{m \times n}$ with rows partitioned into $k$ groups $\groups = \{\bfA_1, \ldots, \bfA_k\}$, find a unit vector $\bfv \in \reals^n$ such that:
\begin{align*}
\min_{\bfv \in \reals^n, \, z \in \reals} & \quad z \\
\text{s.t.} \quad & \ssq(\Ait) - \bfv^{\top}\Ait\bfv \leq z \quad \forall \bfA_i \in \groups, \\
& \|\bfv\|_2^2 = 1.
\end{align*}
\end{problem}

We refer to the left-hand side of the constraints as \emph{constraint functions}, defined for each group $i$ as:
\[
h_i(\bfv) = \ssq(\Ait) - \bfv^{\top}\Ait\bfv.
\]

\spara{Convexity Analysis.} Problem~\ref{prob:faireig_optimization} is non-convex. Each quadratic form $-\bfv^\top \bfA_i^\top \bfA_i \bfv$ is concave since $-\bfA_i^\top \bfA_i$ is negative semidefinite. Hence, each $h_i(\bfv)$ is concave, and the minimization of a maximum over concave functions is a non-convex optimization problem. However, all $h_i$ are continuous over the unit sphere, and each attains a global minimum of zero. Using this insight, we can establish a key optimality condition (proof in Appendix \ref{app:proofs}):

\begin{theorem}
\label{theorem:equal_loss}
Let $(\bfv^*, z^*)$ be an optimal solution to Problem~\ref{prob:faireig_optimization}. Then, there exist distinct groups $i \neq j$ such that:
\[
z^* = h_i(\bfv^*) = h_j(\bfv^*) \geq h_k(\bfv^*) \quad \forall k \notin \{i, j\}.
\]
\end{theorem}

\spara{Two-group case.} As we stated before, Problem \ref{prob:faireig_optimization} is tractable when there are two groups. Theorem~\ref{theorem:equal_loss} implies that the optimum lies at the intersection of two ellipsoids defined by $h_1(\bfv) = h_2(\bfv)$. Geometrically, this suggests that we can start from the leading eigenvector of one group and follow a descent path toward the intersection point. We formalize this intuition using KKT conditions in Section~\ref{sec:twogroup_analysis}, where we show that the two-group case enjoys strong duality. This aligns with known results for problems with two quadratic constraints~\cite[Appendix B]{boyd2004convex}.

\spara{The dual problem.} To better analyze and solve Problem~\ref{prob:faireig_optimization}, we derive its dual, which has a more tractable and informative objective for gradient-based methods such as Frank-Wolfe~\cite{frank1956algorithm}. Notably, even though the primal is non-convex, we prove strong duality when $|\groups| = 2$, and for $|\groups| > 2$, the dual still provides useful bounds on solution quality.

The Lagrangian associated with Problem~\ref{prob:faireig_optimization} is:
\[
\lagrangian(\bfv,z,\bfmu,\lambda)  = z+ \sum_{i=1}^{k} \mu_i (h_i(\bfv) -z) + \lambda ( \|\bfv\|_2^2-1), 
\]
where $\bfmu = [\mu_1, \ldots, \mu_k] \geq 0$ and $\lambda$ are dual variables. Define:
\[
\bfA(\bfmu) = \sum_{i=1}^k \mu_i \bfA_i^\top \bfA_i, \quad \bfs = [\ssq(\bfA_1^{\top}\bfA_1),\ldots,\ssq(\bfA_k^{\top}\bfA_k)].
\]

Then the dual problem becomes:

\begin{problem}[\dual]
\label{prob:dual}
{\begin{align}
\max_{\bfmu \in \reals^k} \, \, & \bfmu^\top\bfs - \lambda_{max}(\bfA(\bfmu)) \nonumber \\
\text { s.t. }  & \label{constr:sum_mu}  \bfone^\top\bfmu =1 \\
                & \bfmu\geq 0. \label{constr:geqzero}
\end{align}} 
\end{problem}
A full derivation of the dual is provided in Appendix~\ref{app:dual}.

Problem~\ref{prob:dual} is convex and admits an intuitive interpretation: the optimal direction $\bfv$ is the leading eigenvector of a convex combination of the group-wise covariance matrices (after centering), weighted by $\bfmu$. This mirrors classical \pca, where the principal component is the leading eigenvector of the global covariance matrix.

\spara{Uniqueness.} Later, we will define the optimal solution $\bfv$ as a function of $\bfmu$. However, $\bfv$ is not always unique, as $\bfA(\bfmu)$ may have repeated eigenvalues. In practice, this is rarely an issue: real-world data typically avoid eigenvalue degeneracies due to noise~\cite{kato1966perturbation}. If needed, slight perturbations can be introduced to ensure uniqueness.

\section{Algorithms for \faireig}
\label{sec:algos}
We present two algorithms for solving the \faireig problem. The first is a scalable, gradient-based method that solves the dual problem (Problem~\ref{prob:dual}) using the Frank-Wolfe algorithm. The second is a semidefinite programming (SDP) relaxation of the primal problem, which can provide more accurate but less scalable solutions.

\spara{Frank-Wolfe.} The Frank-Wolfe algorithm~\cite{pokutta2023frank} is an iterative method for constrained convex optimization. At each iteration, it linearizes the objective and moves toward a solution that maximizes the linear approximation within the feasible set.

This approach is well-suited for Problem~\ref{prob:dual}, as the feasible region is the standard simplex (simplex constraints: $\bfmu \ge 0$, $\bfone^\top \bfmu = 1$), and the objective is differentiable. The primary computational bottleneck is computing the gradient of the dual objective:
\[
g(\bfmu) = \bfmu^\top \bfs - \lambda_{\max}(\bfA(\bfmu)),
\]

Let $\bfv(\bfmu)$ be the leading eigenvector of $\bfA(\bfmu)$, i.e., $\bfA(\bfmu) \bfv(\bfmu) = \lambda(\bfmu) \bfv(\bfmu)$, where $\lambda(\bfmu) = \lambda_{\max}(\bfA(\bfmu))$. By differentiating this eigenvalue equation and using orthogonality of $\bfv(\bfmu)$ and its gradient (via the constraint $\|\bfv(\bfmu)\|_2 = 1$), we obtain the gradient:
\begin{equation}
\label{eq:final_gradient}
(\nabla g)_i = \bfs_i - \bfv(\bfmu)^\top \bfA_i^\top \bfA_i \bfv(\bfmu).
\end{equation}

Algorithm~\ref{alg:FW} summarizes the full Frank-Wolfe procedure for solving the dual. The algorithm iteratively updates the dual variable $\bfmu$ using gradient information and solves a linear subproblem over the simplex at each step. The computational cost is dominated by the eigenvalue computation in Line 5, which can be performed efficiently via the Lanczos method. The overall complexity is $\mathcal{O}(t n^2)$, where $t$ is the number of iterations until convergence

\begin{algorithm}[t!]
\footnotesize
\caption{Frank-Wolfe for \dual}\label{alg:FW}
\begin{algorithmic}[1]
\STATE \textbf{Input:} Matrices 
$\bfA_1,\ldots,\bfA_k$, convergence tolerance $\epsilon$. 
\STATE \textbf{Initialize:} Set $\bfmu^{(0)}=[1,0,\ldots,0]$, \\$\bfs=[\ssq(\bfA_1^{\top}\bfA_1),\ldots,\ssq(\bfA_k^{\top}\bfA_k)]$
\STATE $t \gets 0$
\REPEAT
    
   \STATE \label{code:eigenproblem} $\bfv(\bfmu^{(t)}) \gets \mathbf{x}$ s.t. $\bfA(\bfmu^{(t)})\mathbf{x}=\lambda_{max}\mathbf{x}$
    
    \STATE \label{code:gradient} $\nabla g(\bfmu^{(t)})_i \gets \bfs_i + \bfv(\bfmu^{(t)})^{\top} \Ait \bfv(\bfmu^{(t)})$ 
    \STATE \label{code:simplex} $ \bfs^{(t)} \gets \arg\max_{\bfy: \bfone^{\top}\bfy = 1, \bfy \geq 0} \bfy^{\top} \nabla g(\bfmu^{(t)})$ 
    \STATE \label{code:updgamma} $\gamma_t \gets \frac{2}{t+2}$ 
    \STATE  \label{code:updmu} $\bfmu^{(t+1)} \gets (1 - \gamma_t)\bfmu^{(t)} + \gamma_t \bfs^{(t)}$
    \STATE \label{code:updt} $t \gets t + 1$ 
\UNTIL{$\|\bfmu^{(t)} - \bfmu^{(t-1)}\| < \epsilon$}
\STATE \textbf{return} $\bfmu^{(t)}$, $\bfv(\bfmu^{(t)})$
\end{algorithmic}
\end{algorithm}

Since the dual problem is convex, Algorithm~\ref{alg:FW} converges to the global optimum. However, its objective value provides only a lower bound on the primal objective due to a possible non-zero duality gap, when $|\groups| > 2$.

\textbf{Semidefinite programming.} We also propose a convex relaxation of the primal \faireig problem via semidefinite programming (SDP)~\cite{boyd2004convex}. Although this method has a higher computational cost—typically $\mathcal{O}(n^6)$ for off-the-shelf solvers—it can yield tighter approximations when duality gaps are present. The SDP formulation replaces the rank-$1$ outer product $\bfv \bfv^\top$ with a matrix variable. Details and pseudocode for the SDP relaxation are provided in the appendix (Algorithm~\ref{alg:sdp}). As shown in our experiments, this approach often produces solutions that are close to rank-$1$ and achieve better primal objective values than Frank-Wolfe, albeit at a significantly higher runtime.

\section{Algorithm and Analysis for Two Groups}
\label{sec:twogroup_analysis}

In many practical scenarios, data are divided into exactly two groups—for example, based on binary attributes. In this case, we show that \faireig (Problem~\ref{prob:faireig_optimization}) can be solved optimally and efficiently, with the additional property that the optimal solution equalizes the loss across both groups.

\spara{Algorithm.}
We present a specialized algorithm for the case $|\groups| = 2$, which outperforms generic methods (e.g., Frank-Wolfe) in both speed and accuracy. When there are two groups, the optimal dual variable $\bfmu$ lies on a one-dimensional simplex and satisfies the equal-loss condition (Theorem~\ref{theorem:equal_loss}). As shown in Lemma~\ref{lemma:root_finding}, the optimal value can be efficiently found via root-finding, which we perform using Brent’s method~\cite{brent1971algorithm}.

\spara{Theoretical Analysis.}
The two-group case also admits strong theoretical guarantees. In particular, as implied by Theorem~\ref{theorem:equal_loss}, any optimal solution $\bfv^*$ to Problem~\ref{prob:faireig_optimization} satisfies $h_1(\bfv^*) = h_2(\bfv^*)$. This leads directly to the following lemma (proof in Appendix~\ref{app:proofs}). 

\begin{lemma}
\label{lemma:alg_equal}
For $|\groups| = 2$, Algorithm~\ref{alg:fairsvd} produces an orthonormal set of fair components such that the total incremental error is equal across both groups.
\end{lemma}

Next, we show that \faireig in this setting enjoys strong duality, and hence is efficiently solvable.

\begin{theorem}
\label{theorem:twogroup_duality}
For $|\groups| = 2$, the \faireig problem satisfies strong duality and can be solved optimally using Brent’s method in time $\mathcal{O}(n^2 \log(1/\varepsilon))$, where $\varepsilon$ is the desired accuracy.
\end{theorem}

 \spara{Proof Sketch.} The dual problem (Problem~\ref{prob:dual}) is convex, so its unique optimum can be computed efficiently (e.g., via Frank-Wolfe or Brent’s method). The KKT conditions fully characterize the solution, and strong duality holds: the optimal values of the primal and dual problems coincide. Therefore, solving the dual yields the optimal primal solution as well.

Note that Property~\ref{property:projection} implies that the total time required to obtain a rank-$d$ solution, using Algorithm \ref{alg:fairsvd}, is also polynomial. Finally, a consequence of Theorem \ref{theorem:twogroup_duality} is Lemma \ref{lemma:SDP_tight}, proved in Appendix~\ref{app:proofs}.

\begin{lemma}
\label{lemma:SDP_tight}
For $|\groups| = 2$, the SDP relaxation (Algorithm~\ref{alg:sdp}) is tight, i.e., it recovers a rank-$1$ solution.
\end{lemma}

\section{Experiments}
\label{sec:experiments}

We evaluate our method in both the two-group case—where optimality guarantees hold—and the multi-group case, where such guarantees no longer apply. Nevertheless, we empirically observe that the duality gap remains small across all settings (see Appendix~\ref{sec:additional_experiments}), indicating near-optimal performance in practice. Our results demonstrate that our method offers substantial improvements over existing approaches for \fairpca.

\subsection{Experimental Setup}
\label{sec:settings}

\textbf{Datasets.}  
We rely on real-world datasets with two or more groups that are also used in related works. 

\begin{itemize}[leftmargin=10pt]
    \item \textbf{Datasets with two groups:} we use the juvenile recidivism dataset from Catalunya (\recidivism)~\cite{tolan2019machine}, and several datasets from the UCI repository~\cite{dua2019uci}, including \heart, \german, \credit, \student, \adult, \compas, and \communities. Group membership is based on sex, except for \communities, where groups are defined by racial majority (caucasian or not).
    
    \item \textbf{Datasets with more than two groups:} we partition \compas into three age-based groups (\compasthree) and \communities into four ethnic groups, black, hispanic, asian, and caucasian (\communitiesfour).
\end{itemize}

We pre\-process data by removing protected attributes, applying one-hot encoding to categorical features, and standardizing all columns group-wise. 
The datasets have up to $1,994$ rows and $227$ features, and exhibit markedly different characteristics, e.g., in terms of unbalance in group sizes.  
Table~\ref{table:dataset-stats} in Appendix~\ref{sec:details_datasets} provides more detailed information for all datasets used in the experiments.

\spara{Baselines.}
We compare our method (\fairsvd) to two recent algorithms for \fairpca: \fairpcaalgo, a semidefinite programming (SDP) approach designed by~\citet{tantipongpipat2019multi} and \bicriteria, a bicriteria approximation method introduced by~\citet{song2024socially} (Algorithm 3).

Given a target rank $d$, both baselines produce a rank-$d$ projection matrix $\bfP = \bfU\Lambda\bfU^\top$, where $\bfU \in \reals^{n \times d}$ is computed via SVD. We assess consistency by comparing the solutions given by the top-$r$ vectors from each method ($r < d$).

\spara{Metrics and parameters.}\label{sec:metrics_parameters}
We report three metrics considered in \fairpca:
the marginal loss (optimized by \citet{samadi2018price}), our incremental loss (see Section~\ref{sec:overview}), and the standard $L_2$ reconstruction loss.
The marginal and incremental losses quantify deviations from the optimal reconstruction, unlike the $L_2$ reconstruction loss.
The \bicriteria algorithm targets the $L_2$ loss, ignoring the optimal reconstruction,
and is thus less competitive in terms of marginal or incremental loss.
We also report runtime (in seconds) for all methods.
Regarding the target rank parameter, $d$, we vary it from $1$ to $8$. 

\spara{Implementation and Hardware.}  
All methods are implemented in \texttt{Python}. For two-group settings, \fairsvd uses the root-finding algorithm (Section~\ref{sec:twogroup_analysis}); for more than two groups, it uses the Frank-Wolfe approach. Experiments are executed on a machine with 32 cores and 256GB RAM. The (anonymized) source code and the datasets used in the experiments are publicly available at:  
\url{https://anonymous.4open.science/r/FairPrincipalComponents/}. 

\subsection{Results for Two-group Data}\label{sec:exp_results}

\begin{figure*}[t]
\centering
\begin{minipage}{0.98\textwidth} 
\centering
\includegraphics[width=0.55\textwidth]{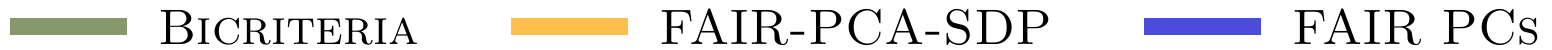} \\
\begin{tabular}{@{}ccc@{}} 
    \includegraphics[width=0.31\textwidth]{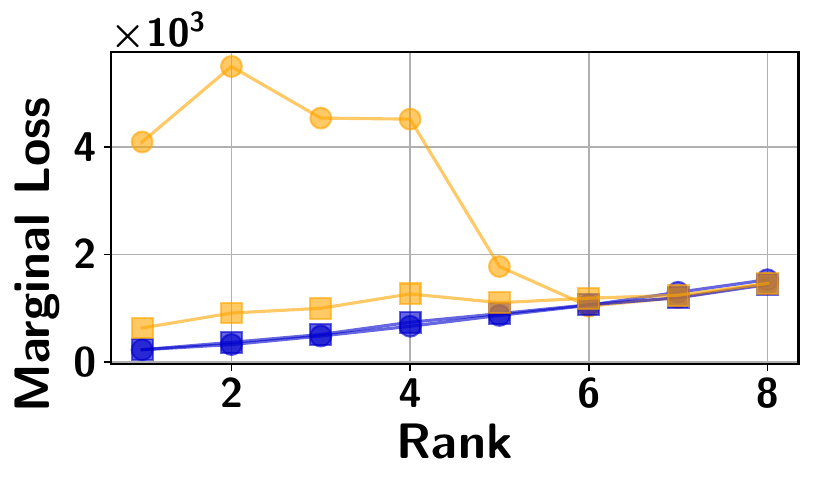} &
    \includegraphics[width=0.31\textwidth]{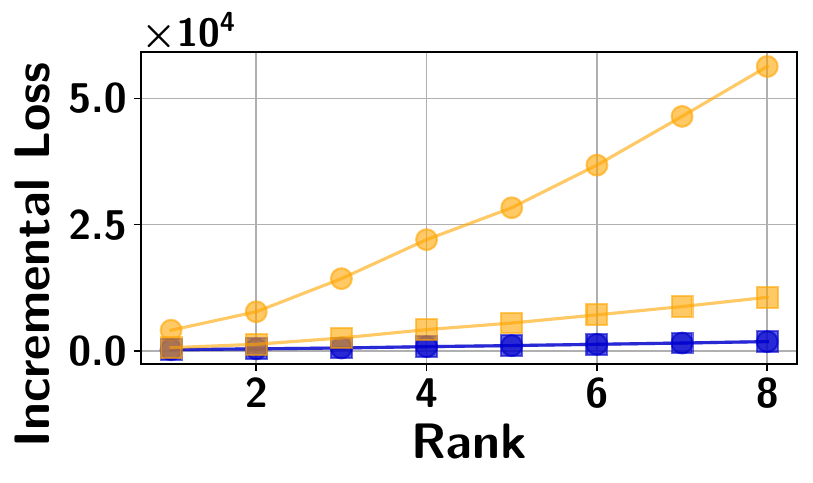} &
    \includegraphics[width=0.31\textwidth]{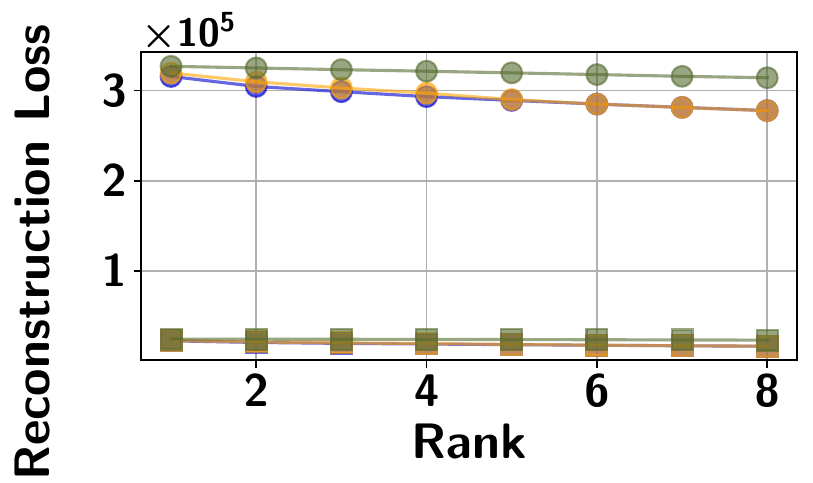} \\
    \includegraphics[width=0.31\textwidth]{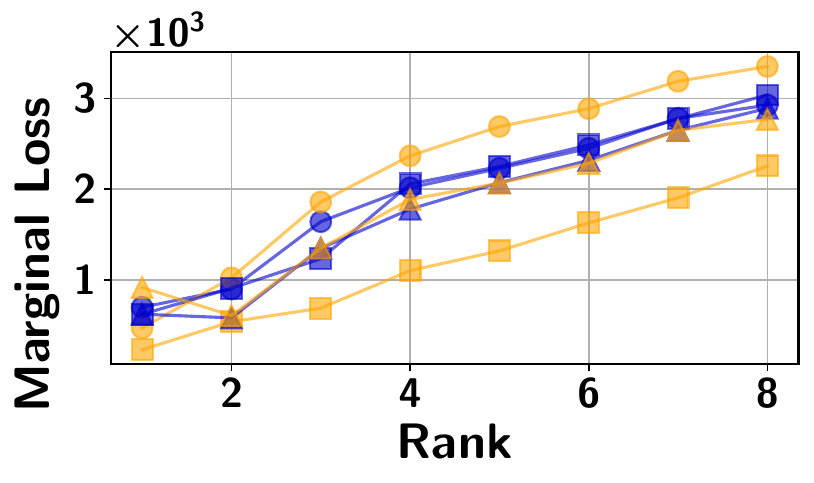} &
    \includegraphics[width=0.31\textwidth]{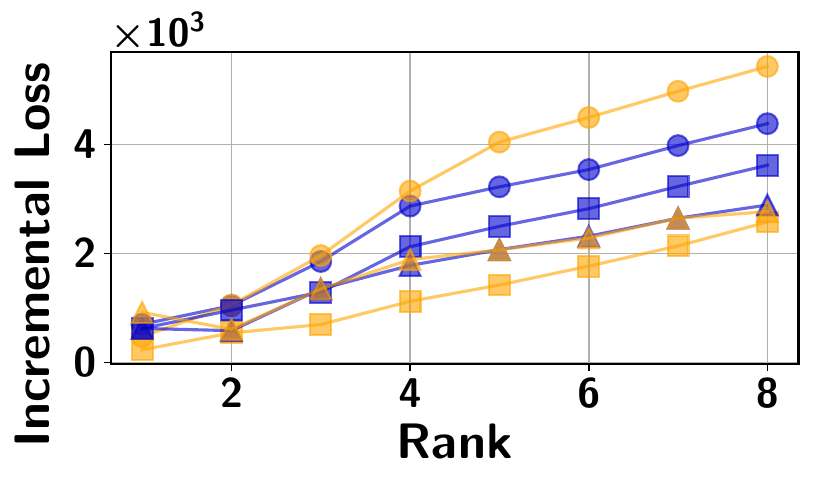} &
    \includegraphics[width=0.31\textwidth]{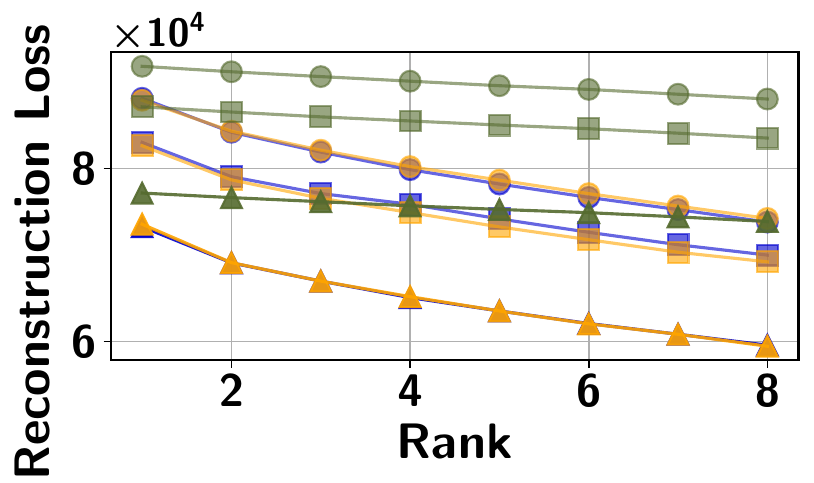} \\
\end{tabular}
\caption{Results on the \compas dataset. Top: two groups. Bottom: three groups. Columns show marginal loss, incremental loss, and $L_2$ reconstruction loss as a function of target rank $d$. Marker symbols indicate different groups.}
\label{fig:peeformance}
\end{minipage}
\end{figure*}

Figure~\ref{fig:peeformance} (top) shows marginal, incremental, and reconstruction loss on the \compas dataset as the target rank increases. Additional results for all other datasets are in the appendix (Figure~\ref{fig:performance_appendix}).

\fairsvd consistently achieves equal incremental loss across both groups for all $d < 8$, preserving fairness in all lower-rank subspaces—a property not satisfied by \fairpcaalgo or \bicriteria. Moreover, the incremental loss for \fairsvd is substantially smaller than that of \fairpcaalgo. Although \fairpcaalgo is optimized for marginal loss, \fairsvd often achieves comparable or better performance on that metric as well. In terms of $L_2$ reconstruction loss, \fairsvd performs similarly to \fairpcaalgo and outperforms \bicriteria, which is designed for that metric.

\spara{Runtime comparison.} 
Table~\ref{tab:runtimes} reports runtime (in seconds) for each method with target rank $d = 8$. \bicriteria is typically the fastest, but it is not competitive in terms of performance. \fairpcaalgo is slow on larger datasets due to the overhead of SDP solvers. In contrast, \fairsvd runs in under 3 seconds across all datasets and scales significantly better than \fairpcaalgo, while achieving consistently competitive or superior performance according to all evaluation criteria.

\begin{table}[t]
\footnotesize
\centering
\setlength{\tabcolsep}{3pt}
\caption{Runtime (in seconds) of each method for $d = 8$.}
\label{tab:runtimes}
\begin{tabular}{lrrr}
\toprule
Dataset & \fairsvd & \fairpcaalgo & \bicriteria \\
\midrule
\heart              & 0.009  & 0.022   & 0.016 \\
\german             & 0.100  & 0.900   & 0.021 \\
\credit             & 0.230  & 0.084   & 0.053 \\
\student            & 0.067  & 0.640   & 0.031 \\
\adult              & 2.160  & 9.130   & 0.200 \\
\compas             & 0.710  & 143.150 & 0.053 \\
\communities        & 0.280  & 8.620   & 0.035 \\
\recidivism         & 1.280  & 357.590 & 0.061 \\
\compasthree        & 2.540  & 124.110 & 0.019 \\
\communitiesfour    & 1.230  & 11.160  & 0.024 \\
\bottomrule
\end{tabular}
\vspace*{-0.2cm}
\end{table}

\subsection{Results for More than Two Groups}\label{sec:mutliple_groups}

When the number of groups exceeds two, 
all algorithms under consideration lose optimality guarantees. However, our experiments suggest that \fairsvd remains an effective heuristic in practice.

Figure~\ref{fig:peeformance} (bottom) presents results on the \compasthree dataset. As the target rank increases, \fairsvd consistently yields more balanced reconstructions across groups compared to \fairpcaalgo and \bicriteria. This suggests that \fairsvd remains a strong choice for \fairpca with $|\mathcal{G}| > 2$.  Analogous results for the \communitiesfour dataset, shown in appendix (Figure~\ref{fig:communities_four_groups}), confirm the same trend. In both cases, neither marginal nor incremental losses are equalized across all groups, which is consistent with the lack of equality guarantee in the case of more than two groups.

\spara{Runtime comparison.} Table~\ref{tab:runtimes} also presents the runtimes for experiments involving more than two groups. 
which confirm the trends observed in the two-group setting.

\spara{Empirical duality gap.}
In the case of more than two groups, our algorithms are heuristic and may not reach the true optimum. To quantify this, we compute the \emph{empirical duality gap}, defined as the difference between the primal and dual objective values: $ |f - g| $ where $f = \max_{i} h_i(\bfv) $ and $g$ is the corresponding dual objective value.
Here, $\bfv$ is the solution from either Algorithm~\ref{alg:FW} or Algorithm~\ref{alg:sdp}. A gap of zero indicates that the primal and dual solutions are jointly optimal.

\begin{figure*}[t]
\centering
\begin{minipage}{0.98\textwidth}
\centering
\includegraphics[width=0.35\textwidth]{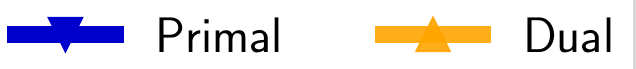} \\
\begin{tabular}{@{}ccc@{}}
    \includegraphics[width=0.31\textwidth]{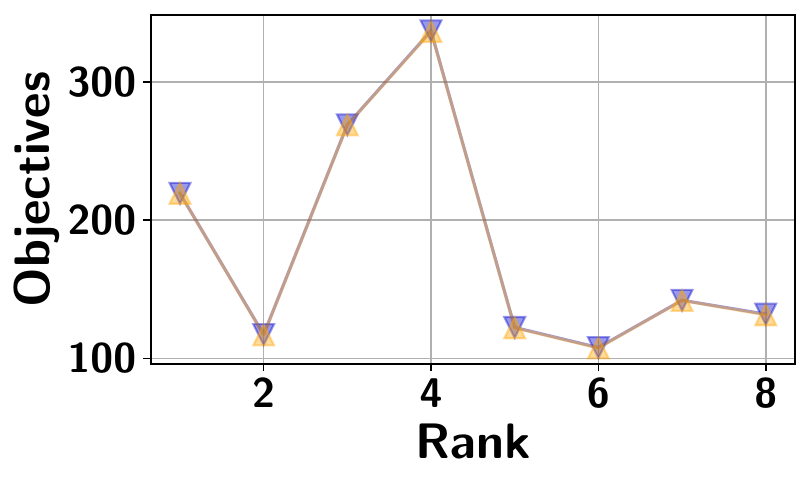} &
    \includegraphics[width=0.31\textwidth]{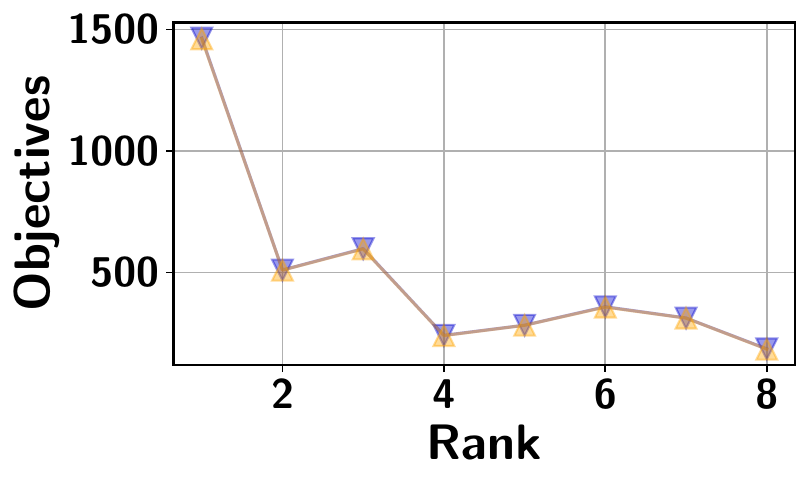} &
    \includegraphics[width=0.31\textwidth]{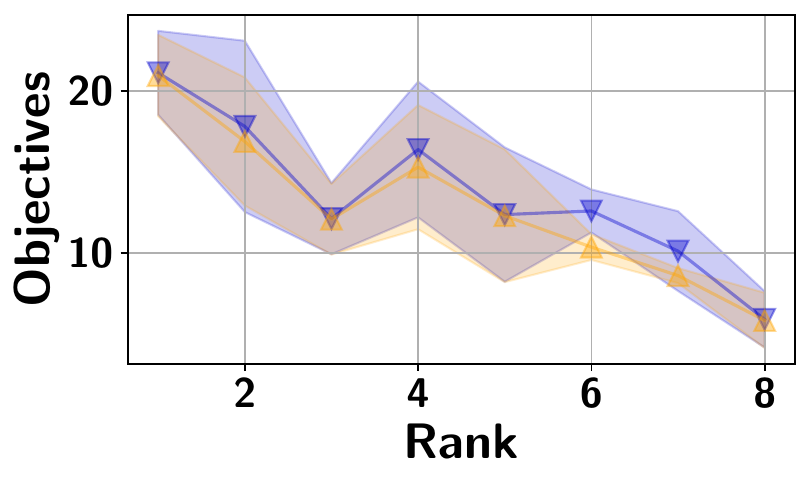} \\
    \compasthree & \communitiesfour & \gaussianthree \\
\end{tabular}
\vspace{-0.1cm}
\caption{Real-world and syntethic data. Primal and dual optimal objective values as a function of rank for the solution relying on the Frank-Wolfe algorithm. 
For  synthetic data (\gaussianthree), the shaded region indicates one standard deviation from the mean across generated datasets.}
\label{fig:primal_and_dual}
\end{minipage}
\end{figure*}

Figure~\ref{fig:primal_and_dual} shows the primal and dual objective values for three datasets: \compasthree (3 groups), \communitiesfour (4 groups), and \gaussianthree (a synthetic dataset with three groups $50 \times 10$ drawn from a Gaussian). Across all datasets, the empirical duality gap is consistently small, demonstrating that \fairsvd closely approximates optimal solutions despite the absence of formal guarantees.

The results in Figure~\ref{fig:primal_and_dual} are obtained using Algorithm~\ref{alg:FW} (Frank-Wolfe), which offers an efficient and scalable approach to solving the dual. While SDP-based solutions (Algorithm~\ref{alg:sdp}) are slower, they tend to yield an even smaller duality gap. This is confirmed in Appendix~\ref{sec:additional_experiments}, Figure~\ref{fig:duality_gap}.

\section{Limitations}
\label{sec:limitations}

Our approach inherits some limitations common to prior work on min-max \fairpca. First, our theoretical guarantees currently apply only in the two-group setting; extending these results to more groups remains an open and important direction for future research. Second, although the min-max formulation is well-established in the fairness literature, it does not ensure parity across all groups when $|\groups| > 2$. Alternative fairness criteria may be necessary in applications where parity is critical. Finally, our method assumes known and fixed group membership; incorporating uncertainty in group labels or supporting intersectional subgroups would broaden applicability in real-world scenarios.

\section{Conclusion}
\label{sec:conclusion}

We introduced a new formulation of Fair PCA that preserves the containment property of standard PCA while minimizing the worst-case reconstruction error across groups, effectively bridging the gap between standard PCA and existing approaches to FAIR PCA. Our method incrementally constructs fair principal components, yielding a sequence of subspaces nested into each other. 

We analyzed the problem of identifying fair principal components, showing it is tractable for two groups and proposing scalable heuristics for the general case.

Empirical results demonstrate that our method can outperform prior work on Fair PCA in both fairness and efficiency. 




\bigskip
\bigskip

\appendix

\section*{Appendix}

\label{appendix}

\section{Derivations related to \textbf{\faireig}-Dual  }\label{app:dual}

In this section, we provide theoretical insights and derivations related to the dual problem associated with  \faireig. 

\subsection{Derivation of the Dual}
 The dual objective is obtained from the Lagrangian as:
\[
g(\bfmu, \lambda)= \inf_{\bfv,z} \lagrangian(\bfv,z,\bfmu,\lambda). 
\]
First, note that in the Lagrangian, the terms involving $z$ appear as:
\[
z \left(1 - \sum_{i \in \groups} \mu_i \right).
\]
Taking the derivative with respect to $z$ and setting it to zero yields:
\[
\frac{\partial \lagrangian}{\partial z} = 0 \quad \Rightarrow \quad \sum_{i \in \groups} \mu_i = 1.
\]
Thus, when this constraint is satisfied, $z$ disappears from the Lagrangian without affecting the optimal solution.

Next, consider the infimum over $\bfv$. Rearranging terms in the Lagrangian, we obtain:
\[
\bfv^\top \left( -\sum_{i \in \groups} \mu_i \Ait + \lambda \matrixI \right) \bfv.
\]
This expression is unbounded below unless the matrix inside the quadratic form is positive semidefinite. Defining: 
\[
\bfA(\bfmu) = \sum_{i \in \groups} \mu_i \Ait,
\]
we require:
\[
-\bfA(\bfmu) + \lambda \matrixI \succeq 0.
\]
Since each $\Ait$ is positive semidefinite and $\bfmu$ is a convex combination (i.e., $\mu_i \geq 0$, $\sum_i \mu_i = 1$), the matrix $\bfA(\bfmu)$ is also positive semidefinite. Its negation is negative semidefinite, so the above constraint is satisfied when $\lambda \geq \lambda_{\max}(\bfA(\bfmu))$.

To tighten the dual bound, we choose the smallest such $\lambda$, i.e., $\lambda = \lambda_{\max}(\bfA(\bfmu))$. Let $\bfs = [\ssq(\bfA_1^{\top}\bfA_1), \ldots, \ssq(\bfA_k^{\top}\bfA_k)]$ denote the vector of group-specific top eigenvalues. The resulting dual problem becomes:
\begin{align}
\max_{\bfmu \in \reals^k} \quad & \bfmu^\top \bfs - \lambda_{\max}(\bfA(\bfmu)) \nonumber \\
\text{s.t.} \quad & \bfone^\top \bfmu = 1 \label{constr:sum_mu2} \\
                 & \bfmu \geq 0. \label{constr:geqzero2}
\end{align}

\subsection{Gradient of the Dual Objective}

Denoting for brevity 
$\lambda(\bfmu) = \lambda_{max}(\bfA(\bfmu))$, 
we have that $\lambda(\bfmu)$ is an eigenvalue of $\bfA(\bfmu)$ and hence: 
\begin{equation}
\label{eq:param_eig} 
\bfA(\bfmu)\bfv(\bfmu)=\lambda(\bfmu)\bfv(\bfmu), 
\end{equation}
where $\bfv(\bfmu)$ is the eigenvector corresponding to $\lambda(\bfmu)$.
Taking the gradient and using the product rule, we have:
\begin{equation}
\label{eq:param_eig_grad}  
\Ait\bfv(\bfmu)+\bfA(\bfmu)\nabla\bfv(\bfmu)=\nabla\lambda(\bfmu)\bfv(\bfmu)+\lambda(\bfmu)\nabla\bfv(\bfmu). 
\end{equation}

To simplify the gradient, we use the constraint $\bfv(\bfmu)^{\top}\bfv(\bfmu)=1$. This gives: 
\[ \nabla\bfv(\bfmu)^{\top}\bfv(\bfmu)+\bfv(\bfmu)\nabla\bfv(\bfmu)=0,
\]
i.e., $\bfv(\bfmu)$ is orthogonal to its gradient. Therefore, multiplying equation \ref{eq:param_eig_grad} with $\bfv(\bfmu)^{\top}$, we obtain:
\begin{align*}
   &\bfv(\bfmu)^{\top}\Ait\bfv(\bfmu)+\lambda(\bfmu)\bfv(\bfmu)^{\top}\nabla\bfv(\bfmu)\\
  = \,  &\nabla \lambda(\bfmu) \bfv(\bfmu)^{\top}\bfv(\bfmu)+\lambda(\bfmu)\bfv(\bfmu)^{\top}\nabla\bfv(\bfmu),
\end{align*}
which simplifies to $(\nabla\lambda(\bfmu))_i=\bfv(\bfmu)^{\top}\Ait\bfv(\bfmu)$.
Putting everything together, we conclude that: 
\begin{equation}
\label{eq:final_gradient}
(\nabla g)_i = \bfs_i - \bfv(\bfmu)^{\top}\Ait\bfv(\bfmu). 
\end{equation}

\section{SDP}

Algorithm \ref{alg:sdp} contains the pseudocode of \sdp to solve Problem~\ref{prob:faireig_optimization}. 

\begin{algorithm}
\caption{\faireig-SDP}
\begin{algorithmic}[1]
\label{alg:sdp}
\STATE \textbf{Input:} Matrices $[\bfA^1,\ldots,\bfA^k]$
\STATE $\matrixX\in \mathbb{R}^{n\times n}\gets$ \textbf{Solve}: 
\begin{align}
\min\nolimits_{\text{ }z\in\reals}& \ \ \ z \label{sdp}\\
\text { s.t. } & \ssq(\Ait) - \text{Tr}(\bfA^i\matrixX) \leq z \quad \text{for} \quad \bfA^i \in \groups \nonumber \\
&  \begin{bmatrix} \matrixX & \bfx \\ \bfx^\top & 1 \end{bmatrix} \succeq 0 \nonumber \quad \text{,Tr}(\matrixX) \leq 1 ,  \nonumber
\end{align} 
\STATE $\matrixX= \sum_{j=1}^n {\lambda}_j\bfx_j \bfx_j^\top$
\STATE \textbf{Output:} $\bfx_1 \in \reals^{n}$
\end{algorithmic}
\end{algorithm}

\section{Loss Functions}
\label{app:loss_functions}

In this section we justify our choice of loss function and contrast it with alternative formulations.

Before we proceed we define some additional notation.
\subsection{Notation}
For $\bfV \in\reals^{n\times d}$, we write $\{\bfV\} = \{\bfv_1, \ldots, \bfv_d\}$ to denote its ordered columns. We denote the orthogonal complement of the span of $\bfV\in\reals^{n\times d}$ by $\Vc$.  
Let $\bfV_{:r} \in \reals^{n \times r}$ denote the matrix containing the first $r$ columns of $\bfV$. 
Given a matrix $\bfV \in \reals^{n \times d}$ with orthonormal columns,  
the projection of $\bfA$ onto $\bfV_r^{\perp}$ is given by $\bfA - \bfA \bfV_r \bfV_r^\top$.

\subsection{Choice of loss function}
We define the our loss function as:
\[
\loss(\matrixM, \bfv) = \max_{\bfA_i \in \groups} \left\{ \ssq(\Ait) - \bfv^\top \Ait \bfv \right\},
\]
where $\ssq(\Ait)$ is the maximum variance achievable by any rank-$1$ projection of group $\bfA_i$, and $\bfv^\top \Ait \bfv$ is the variance captured by the direction $\bfv$.

An alternative might be to maximize the minimum variance captured across groups:
\[
\mathcal{P}(\matrixM, \bfv) = \min_{\bfA_i \in \groups} \left\{ \bfv^\top \Ait \bfv \right\}.
\]

While $\mathcal{P}$ is a natural objective for fairness, it exhibits problematic behavior in multi-group min-max settings. Consider two groups $\matrixM = \{\bfA, \bfB\}$ with corresponding top eigenvectors $\bfw_A$, $\bfw_B$, and assume $\bfw_A^\top \A^\top \A \bfw_A = \ell_A \gg \ell_B = \bfw_B^\top \B^\top \B \bfw_B$. In this case, $\mathcal{P}$ is upper-bounded by $\ell_B$, regardless of how much variance $\bfv$ can capture for group $\bfA$.
This may be considered unfair to group $A$, since it suffers from a poor reconstruction only due to the fact that group $B$ cannot be represented as well by a $1-d$ line. 

\subsection{Marginal Loss}

To address this issue, \citet{samadi2018price} proposed the \emph{marginal loss}, defined next. 
\begin{definition}[Marginal loss]
Given a group matrix $\bfA_g$ and a projection matrix $\bfV \in \VV_d$, the marginal loss is
\[
\loss_{\text{marg}}(\bfA_g, \bfV) \triangleq \|\bfA_g^d - \bfA_g \bfV \bfV^\top\|_F^2,
\]
where $\bfA_g^d$ denotes the best rank-$d$ approximation of $\bfA_g$.
\end{definition}
We refer to~\citet{samadi2018price, tantipongpipat2019multi} for more background on this loss.

\subsection{Containment vs. Parity}

A desirable property of the marginal loss is that, under certain conditions, it leads to equal group loss in the two-group setting (see Theorem 4.5 in~\citet{samadi2018price}). However, when requiring containment—i.e., that the rank-$d$ subspace contains all lower-rank solutions—parity in marginal or reconstruction loss may no longer hold.

To illustrate this, assume Algorithm~\ref{alg:fairsvd} is run on two groups, $\bfA \in \reals^{a \times n}$ and $\bfB \in \reals^{b \times n}$, and that the loss is either reconstruction or marginal loss. Let $\bfV^*$ be the resulting basis of rank $d$. Then, in general:
\[
\loss(\bfA, \bfV^* \bfV^{*\top}) \neq \loss(\bfB, \bfV^* \bfV^{*\top}).
\]

\paragraph{Case 1: Reconstruction Loss.}
Assume $\loss(\bfA, \bfV_{:d}) = \loss(\bfB, \bfV_{:d})$. Let $\bfv_{d+1} \in \bfV_{:d}^\perp$, and let $\bfA_{d+1}$ be the component of $\bfA$ in $\bfV_{:d}^\perp$. If: 
\[
\|\bfA\|_F^2 - \|\bfA_{d+1} \bfx \bfx^\top\|_F^2 < \|\bfB\|_F^2 - \sigma_1^2(\bfB_{d+1}) \quad \forall \|\bfx\|_2 = 1,
\]
then the losses for the two groups will diverge at rank $d+1$.

\paragraph{Case 2: Marginal Loss.}
Suppose again that $\loss(\bfA, \bfV_{:d}) = \loss(\bfB, \bfV_{:d})$, and that we are selecting $\bfv_{d+1} \in \bfV_{:d}^\perp$. From Property~\ref{property:projection}, parity at rank $d+1$ requires:
\[
\sum_{i=1}^{d+1} \left( \sigma_i^2(\bfA) - \|\bfA \bfv_i \bfv_i^\top\|_F^2 \right) = \sum_{i=1}^{d+1} \left( \sigma_i^2(\bfB) - \|\bfB \bfv_i \bfv_i^\top\|_F^2 \right).
\]
Since equality holds for the first $d$ terms by hypothesis, it must also hold for $i = d+1$. However, if:
\[
\sigma_{d+1}^2(\bfA) - \|\bfA_{d+1} \bfx \bfx^\top\|_F^2 < \sigma_{d+1}^2(\bfB) - \sigma_1^2(\bfB_{d+1}) \quad \forall \|\bfx\|_2 = 1,
\]
then the marginal loss for the two groups will again differ.

These examples highlight that enforcing fairness under marginal or reconstruction loss is not compatible with the containment constraint. Incremental loss (Section \ref{sec:overview}) avoids this issue by design, enabling consistent subspace construction without sacrificing fairness guarantees.

\section{Dataset Details}\label{sec:additional_experiments}\label{sec:details_datasets}

Table~\ref{table:dataset-stats} reports summary descriptive statistics for all the benchmark real-world datasets used in the experiments. Specifically, the table reports the number of features, the number of groups and their sizes (i.e., the number of rows in each group) and the ranks of the matrices associated with each group. 

\begin{table}[t]
\footnotesize
\setlength{\tabcolsep}{4pt}
\caption{Dataset statistics. We report the number of features ($n$), number of groups ($|\mathcal{G}|$), and group-wise row counts and matrix ranks.}
\label{table:dataset-stats}
\centering
\begin{tabular}{lrrrr}
\toprule
Dataset & Features ($n$) & Groups ($|\mathcal{G}|$) & Group Rows & Group Ranks \\
\midrule
\heart              & 14   & 2 & 201, 96        & 13, 13 \\
\german             & 63   & 2 & 690, 310       & 49, 47 \\
\credit             & 25   & 2 & 18,112, 11,888 & 24, 24 \\
\student            & 58   & 2 & 383, 266       & 42, 42 \\
\adult              & 109  & 2 & 21,790, 10,771 & 98, 98 \\
\compas             & 189  & 2 & 619, 100       & 165, 71 \\
\communities        & 104  & 2 & 1,685, 309     & 101, 101 \\
\recidivism         & 227  & 2 & 1,923, 310     & 175, 113 \\
\compasthree        & 189  & 3 & 241, 240, 238  & 115, 110, 97 \\
\communitiesfour    & 104  & 4 & 90, 1,571, 218, 115 & 90, 99, 103, 103 \\
\bottomrule
\end{tabular}
\end{table}

\section{Additional Experiment Results}\label{sec:additional_experiments}

In this section, we present additional experiments.

\subsection{Two Groups}
\begin{figure*}[t]
\centering  
\hspace*{0.3cm}\includegraphics[width=0.65\textwidth]{legend_nips.pdf}\\
\begin{tabular}{ccc}
\hspace*{-0.3cm}&  \heart & \hspace*{-0.3cm} \\[-0.5ex] 
\hspace*{-0.3cm}\includegraphics[height=0.11\textheight]{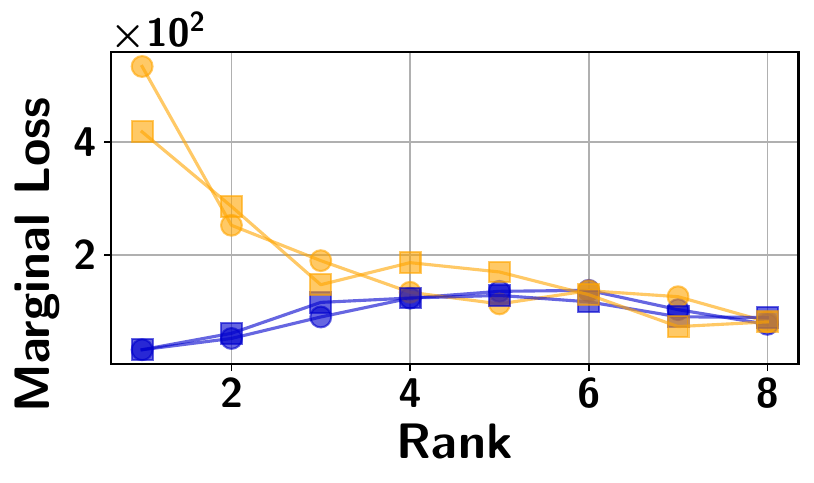} & \hspace*{-0.3cm}\includegraphics[height=0.11\textheight]{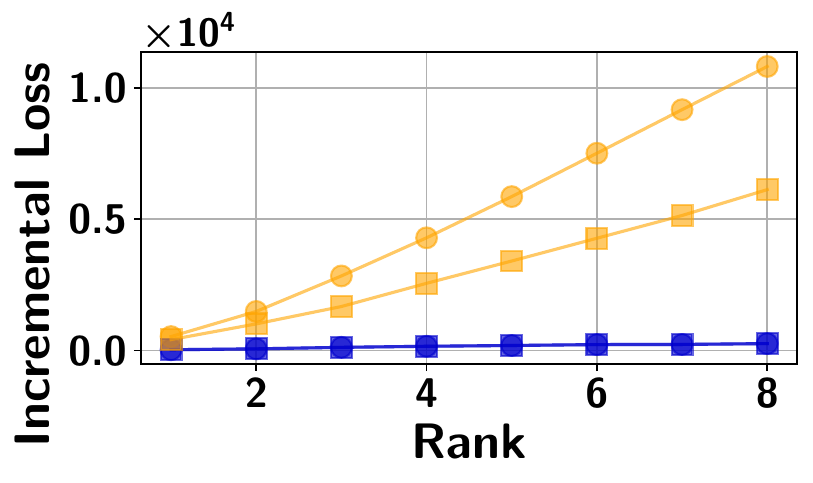} & \hspace*{-0.3cm}\includegraphics[height=0.11\textheight]{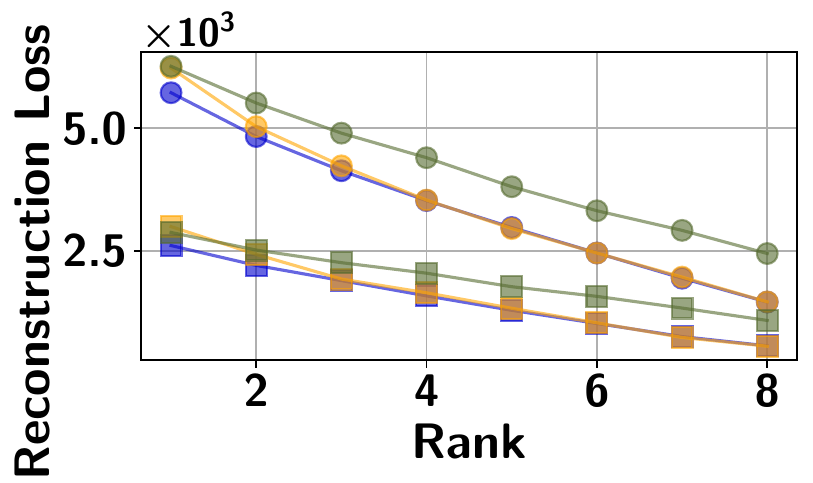} \\
\hspace*{-0.3cm}&  \german & \hspace*{-0.3cm}\\[-0.5ex] 
\hspace*{-0.3cm}\includegraphics[height=0.11\textheight]{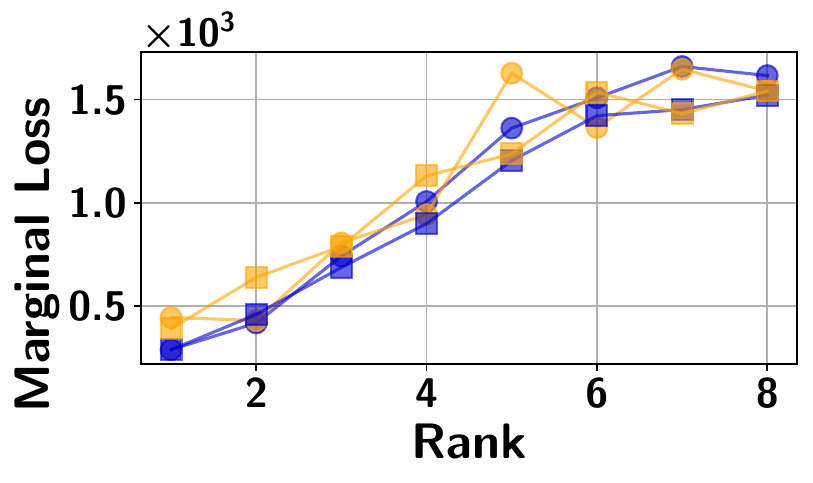} & \hspace*{-0.3cm}\includegraphics[height=0.11\textheight]{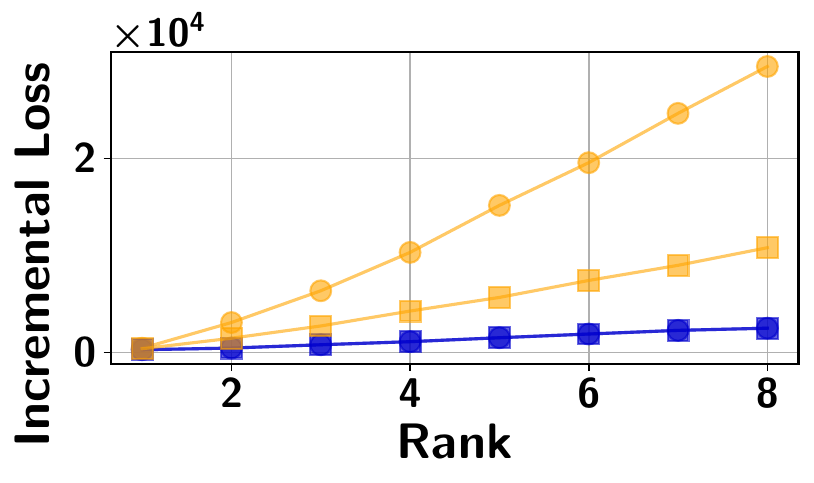} & \hspace*{-0.3cm}\includegraphics[height=0.11\textheight]{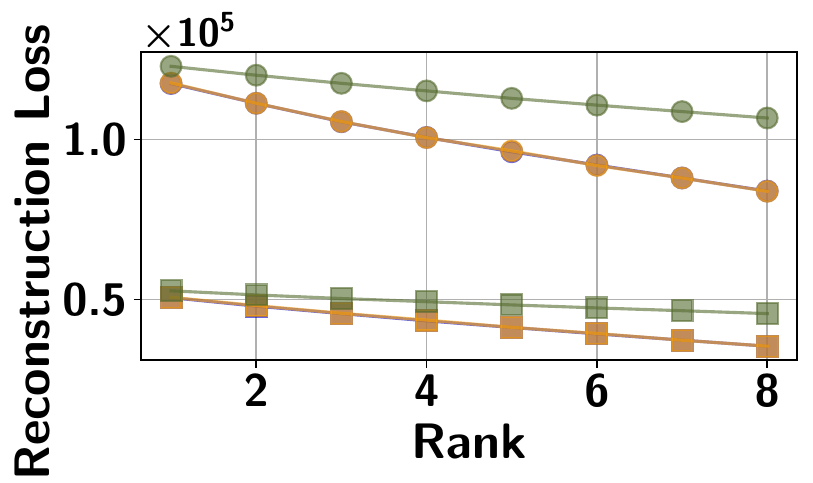} \\
\hspace*{-0.3cm}&  \credit & \hspace*{-0.3cm}\\[-0.05ex]  
\hspace*{-0.3cm}\includegraphics[height=0.11\textheight]{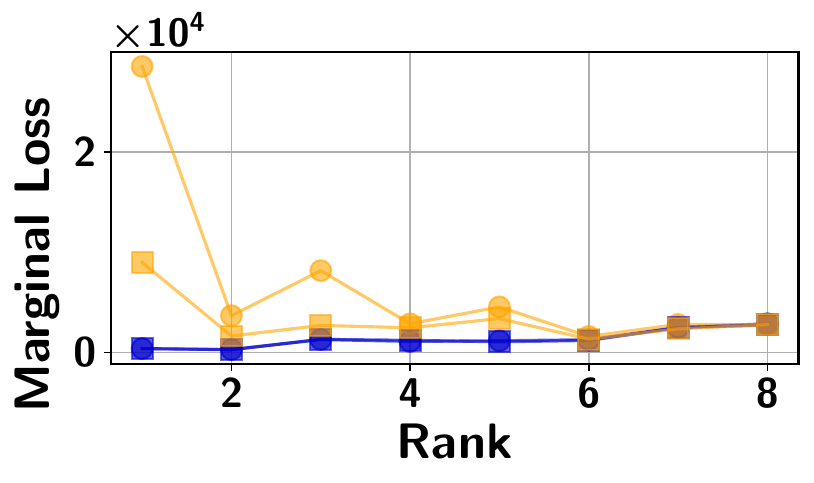} & \hspace*{-0.3cm}\includegraphics[height=0.11\textheight]{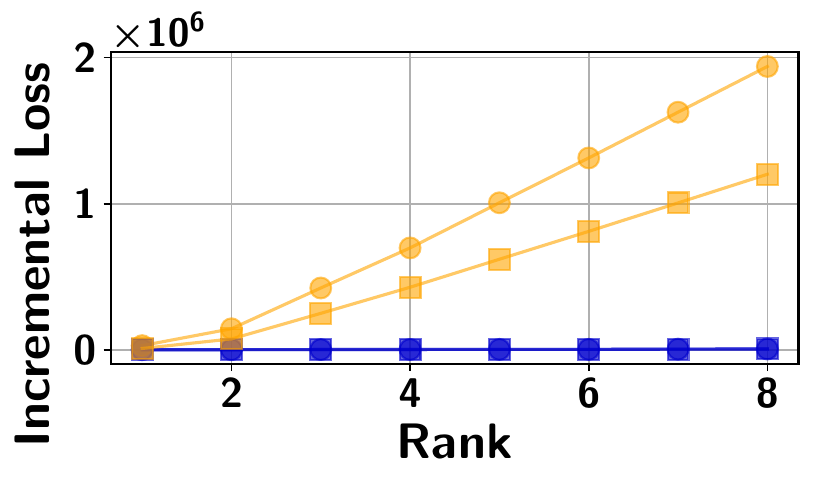} & \hspace*{-0.3cm}\includegraphics[height=0.11\textheight]{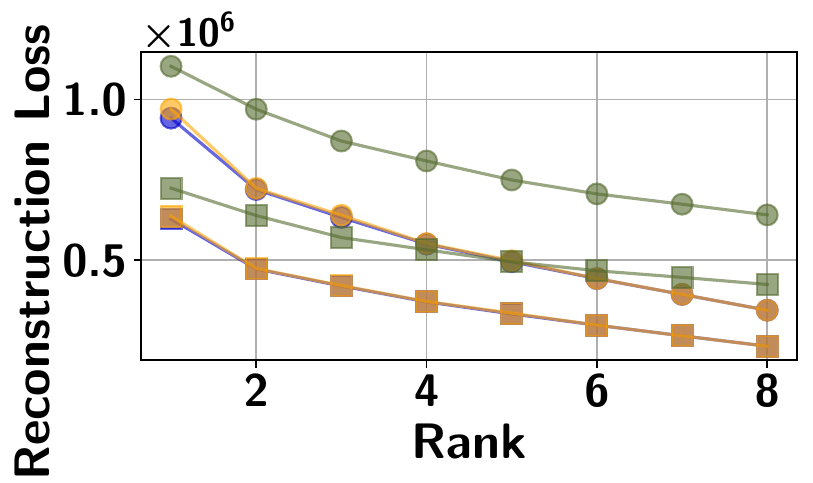} \\
\hspace*{-0.3cm}& \student & \hspace*{-0.3cm}\\[-0.05ex]        
\hspace*{-0.3cm}\includegraphics[height=0.11\textheight]{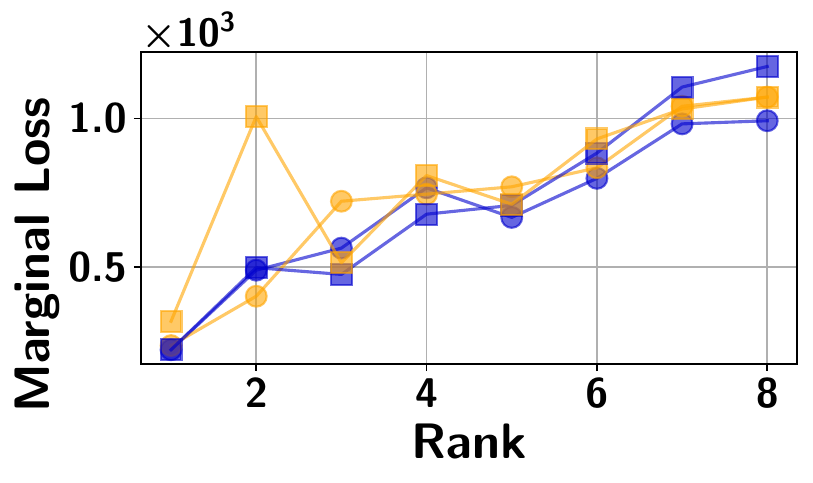} & \hspace*{-0.3cm}\includegraphics[height=0.11\textheight]{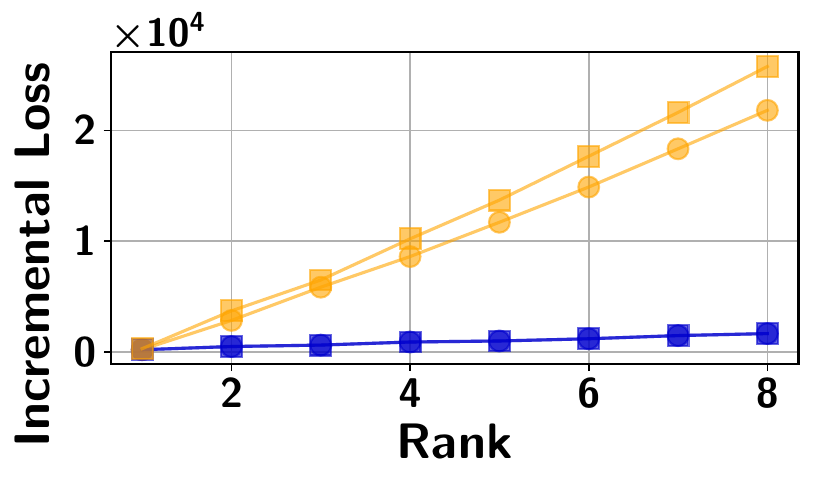} & \hspace*{-0.3cm}\includegraphics[height=0.11\textheight]{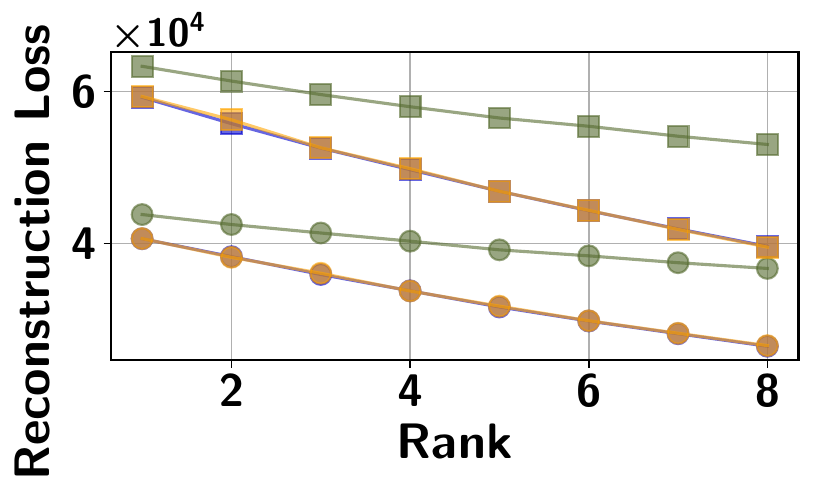} \\
\hspace*{-0.3cm}&  \adult & \hspace*{-0.3cm}\\[-0.05ex]
\hspace*{-0.3cm}\includegraphics[height=0.11\textheight]{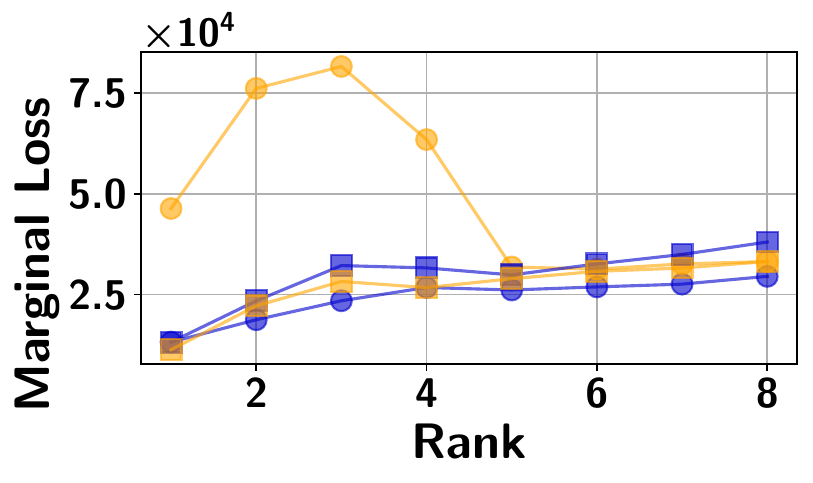} & \hspace*{-0.3cm}\includegraphics[height=0.11\textheight]{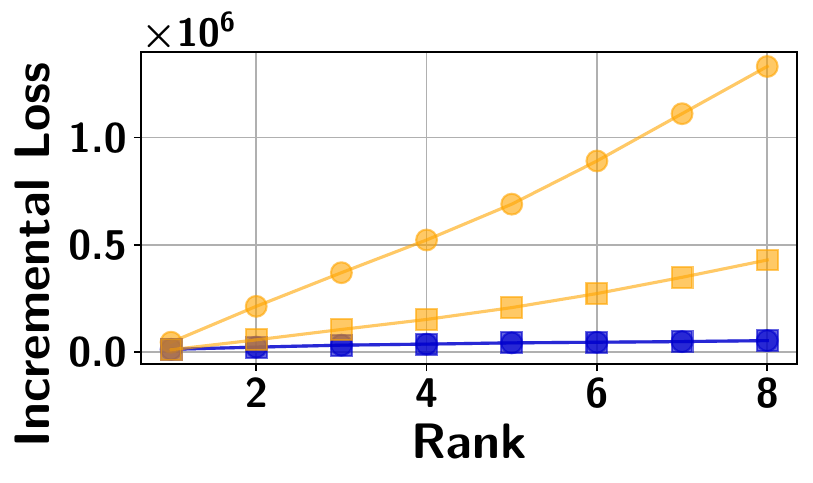} & \hspace*{-0.3cm}\includegraphics[height=0.11\textheight]{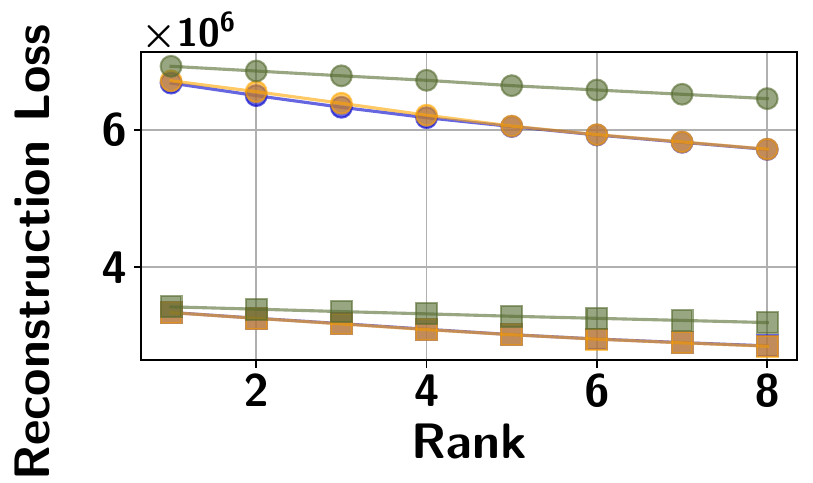} \\
\hspace*{-0.3cm}&  \communities & \hspace*{-0.3cm}\\[-0.05ex] 
\hspace*{-0.3cm}\includegraphics[height=0.11\textheight]{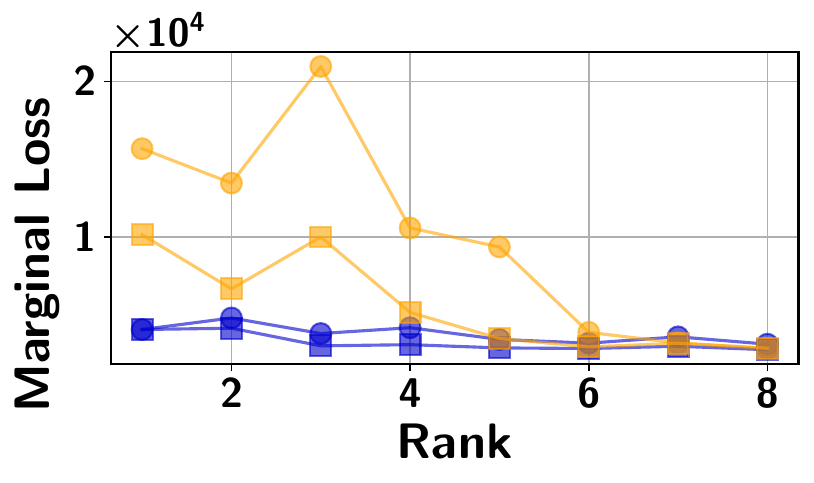} & \hspace*{-0.3cm}\includegraphics[height=0.11\textheight]{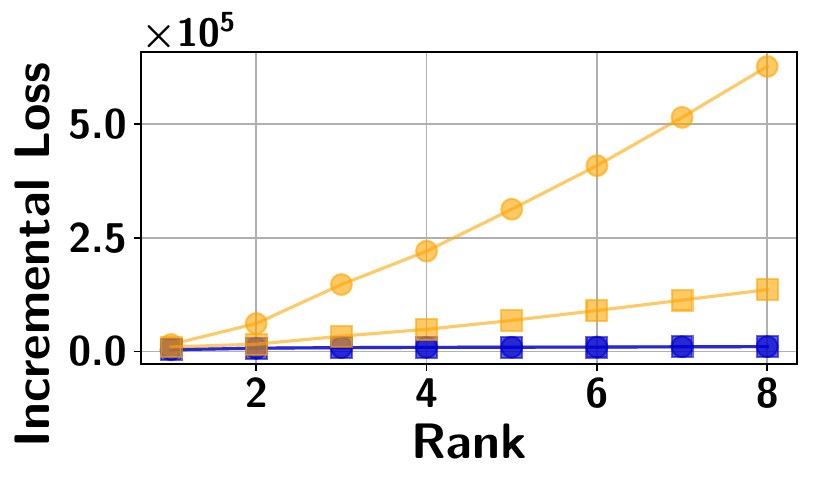} & \hspace*{-0.3cm}\includegraphics[height=0.11\textheight]{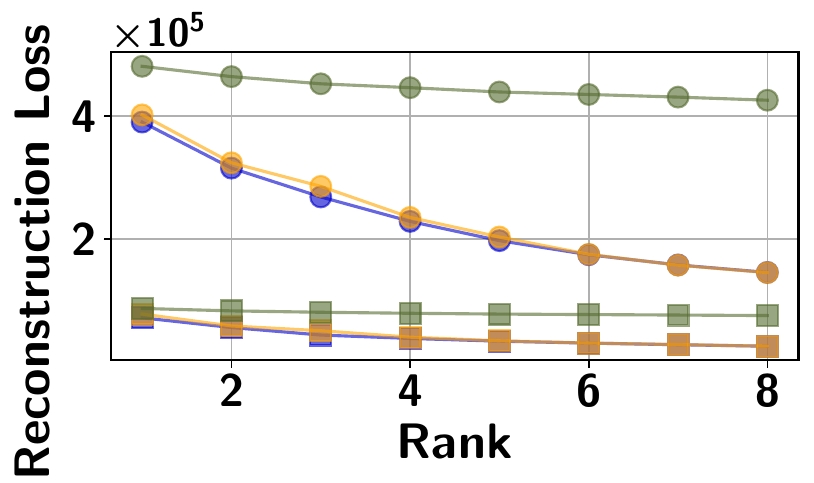} \\
\hspace*{-0.3cm}&  \recidivism & \hspace*{-0.3cm}\\[-0.05ex]   
\hspace*{-0.3cm}\includegraphics[height=0.11\textheight]{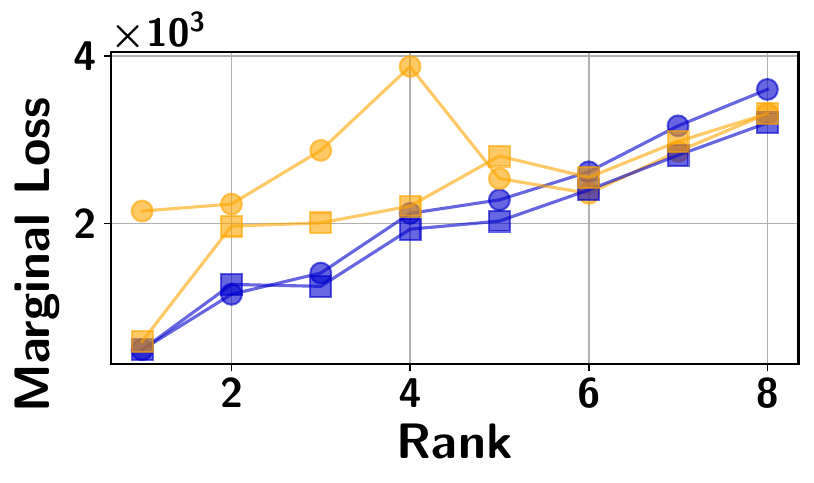} & \hspace*{-0.3cm}\includegraphics[height=0.11\textheight]{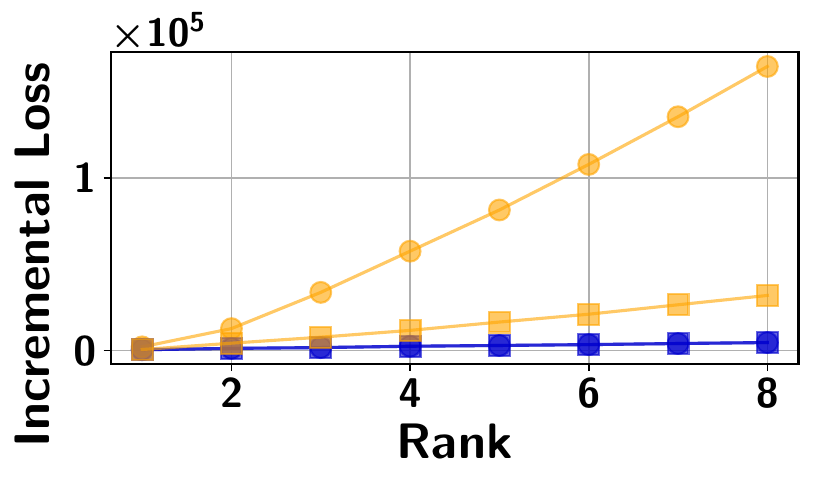} & \hspace*{-0.3cm}\includegraphics[height=0.11\textheight]{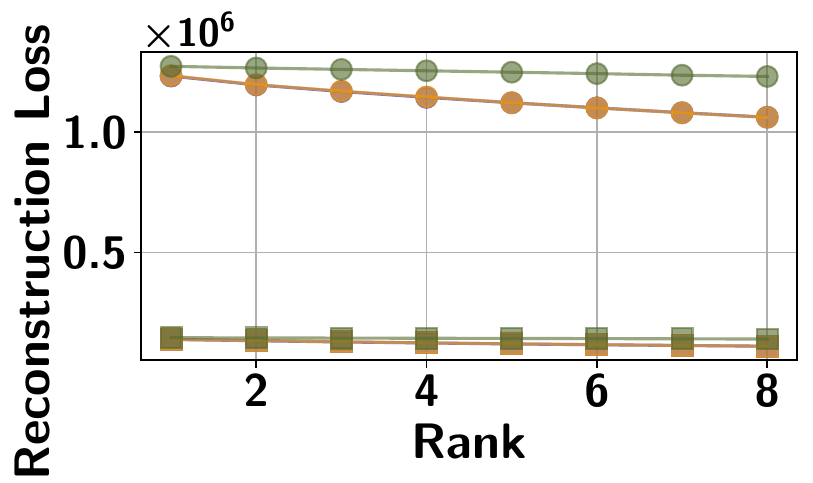} 
\end{tabular}
\caption{Real-world datasets with two groups. Marginal, incremental, and reconstruction loss by rank. Different marker symbols indicate different groups.}
\label{fig:performance_appendix} 
\end{figure*}

Figure~\ref{fig:performance_appendix} shows the different metrics being monitored in our experiments (i.e., the marginal loss, the incremental loss and the reconstruction loss) as a function of reconstruction (target) rank in all considered two-group datasets except the \compas dataset, for which results are provided in Figure~\ref{fig:peeformance} in the main text. 

The findings of the experiments presented in Figures~\ref{fig:performance_appendix} largely corroborate the findings presented in the main text (Figure~\ref{fig:peeformance}) for the \compas dataset. 

\subsection{More than Two Groups}
\begin{figure*}[t]
\centering
\hspace*{0.3cm}\includegraphics[width=0.65\textwidth]{legend_nips.pdf}\\
\begin{tabular}{ccc}
 \hspace*{-0.3cm}\includegraphics[height=0.1275\textheight]{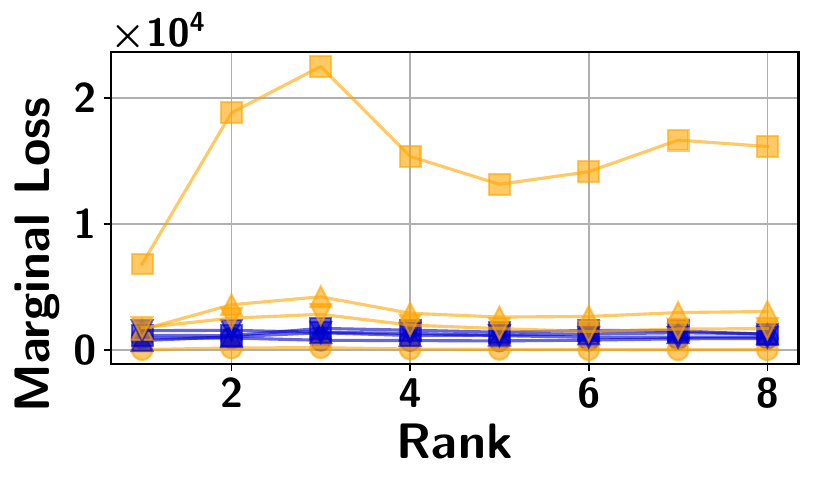} &  \hspace*{-0.3cm}\includegraphics[height=0.1275\textheight]{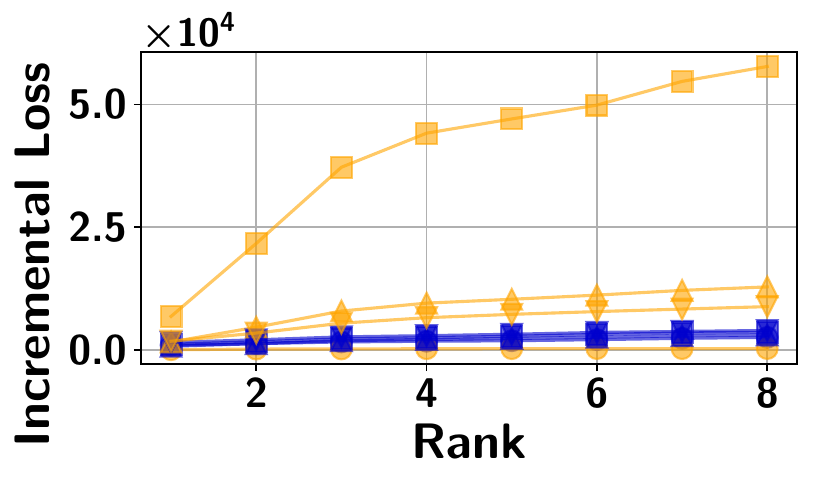} &  \hspace*{-0.3cm}\includegraphics[height=0.1275\textheight]{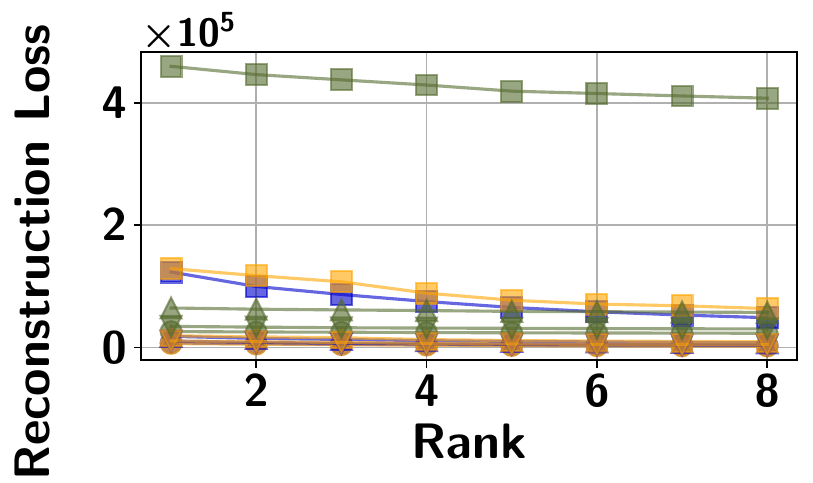} \\
\end{tabular}
\caption{\communitiesfour dataset with four groups. Marginal, incremental and reconstruction loss by rank. 
Different marker symbols indicate different groups.}
\label{fig:communities_four_groups}
\end{figure*}

Figure~\ref{fig:communities_four_groups} displays marginal, incremental and reconstruction loss by rank in the \communitiesfour dataset partitioned into four groups. 
Again, the results for the \communitiesfour dataset are consistent with and confirm the results seen in in Figure~\ref{fig:peeformance} for the \compasthree dataset. 

\begin{figure*}[t]
\centering
\hspace*{0.3cm}\includegraphics[width=0.225\textwidth]{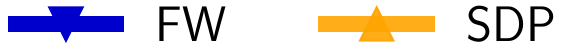}\\
\begin{tabular}{ccc}
 \hspace*{-0.3cm}\includegraphics[height=0.1275\textheight]{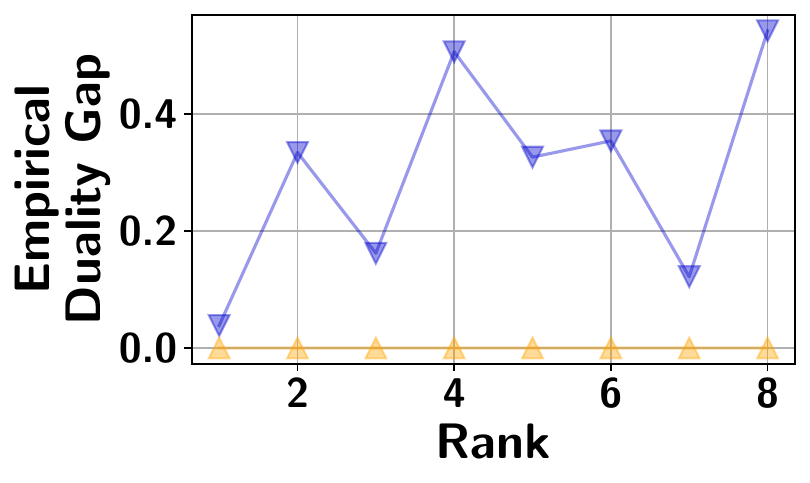} &  \hspace*{-0.3cm}\includegraphics[height=0.1275\textheight]{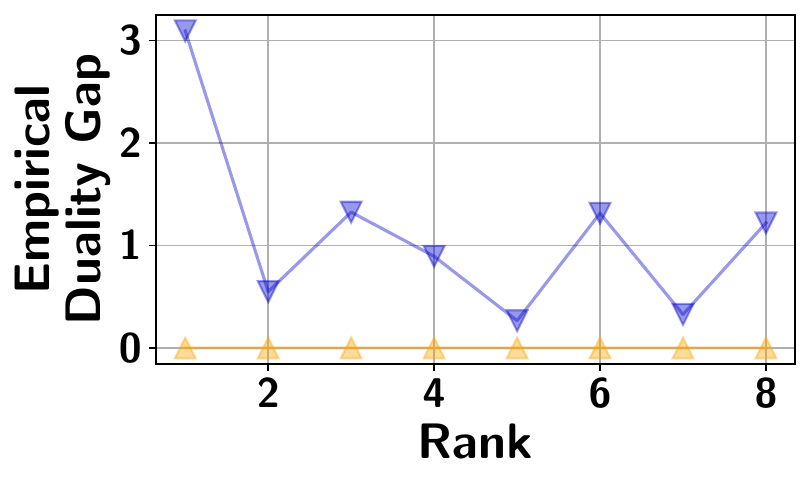} &  
 \hspace*{-0.3cm}\includegraphics[height=0.1275\textheight]{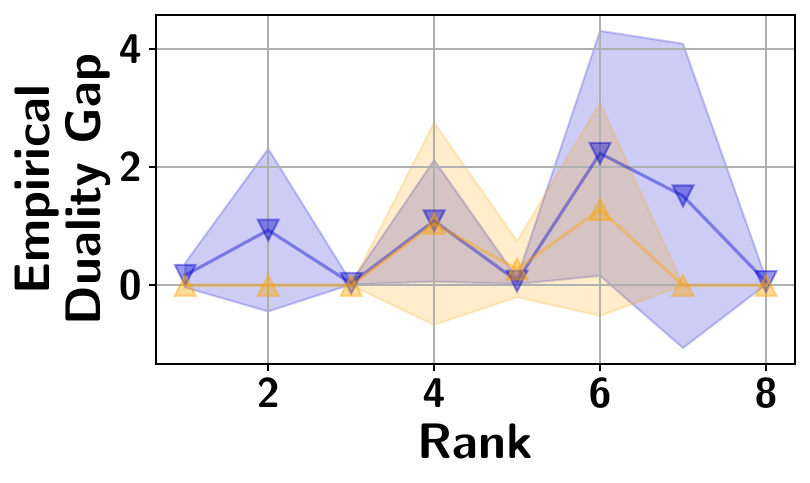} \\
  \hspace*{0.4cm}\compasthree & \hspace*{0.4cm}\communitiesfour & \hspace*{0.4cm}\gaussianthree \\
\end{tabular}
\caption{Real-world and synthetic data. Duality gap as a function of rank for the solutions relying on the Frank-Wolfe (FW) and semidefinite programming solver (SDP). For  synthetic data (\gaussianthree), the shaded region indicates one standard deviation from the mean across generated datasets.}
\label{fig:duality_gap}
\end{figure*}

Finally, Figure~\ref{fig:duality_gap}~shows the empirical duality gap for the proposed solutions based on semidefinite programming and on the Frank-Wolfe algorithm, demonstrating that the formulated semidefinite program, while more time-consuming, can achieve even smaller duality gap than the more efficient approach based on the Frank-Wolfe algorithm.



\section{Proofs}
\label{app:proofs}

All the proofs of our results omitted from the main text 
are detailed in this section.

\subsection{Proof of Property \ref{property:projection}}

\begin{proof}
We have:
\[
\|\bfA \bfV \bfV^\top\|_F^2 = \left\| \sum_{i=1}^d \bfA \bfv_i \bfv_i^\top \right\|_F^2.
\]
Since the vectors $\{\bfv_1, \ldots, \bfv_d\}$ are orthonormal, the projection matrices $\bfv_i \bfv_i^\top$ are pairwise orthogonal. Therefore, the cross terms vanish and:
\[
\left\| \sum_{i=1}^d \bfA \bfv_i \bfv_i^\top \right\|_F^2 = \sum_{i=1}^d \| \bfA \bfv_i \bfv_i^\top \|_F^2.
\]
This concludes the proof.
\end{proof}

\subsection{Proof of Orthonormalization Argument}

\begin{proof}
We prove by induction that the set $\bfV = \{ \bfv_1, \ldots, \bfv_d \}$ constructed by Algorithm~\ref{alg:fairsvd} is an orthonormal set.

\paragraph{Base case ($d=1$).} 
We select an arbitrary unit vector $\bfv_1$. Since it has norm 1, the singleton set $\{ \bfv_1 \}$ forms an orthonormal basis of a one-dimensional subspace.

\paragraph{Inductive hypothesis.} 
Assume that after $k-1$ steps, the vectors $\{ \bfv_1, \ldots, \bfv_{k-1} \}$ form an orthonormal set.

\paragraph{Inductive step.} 
At step $k$, we select $\bfv_k$ from the orthogonal complement of the span of $\{ \bfv_1, \ldots, \bfv_{k-1} \}$. By construction, $\bfv_k$ is orthogonal to all previous vectors, and since it is normalized, the extended set $\{ \bfv_1, \ldots, \bfv_k \}$ remains orthonormal.

By induction, the full set $\bfV = \{ \bfv_1, \ldots, \bfv_d \}$ is orthonormal.
\end{proof}



\subsection{Proof of Theorem \ref{theorem:equal_loss}}

\begin{proof}
Assume for contradiction that $z^* = h_i(\bfv^*) > h_j(\bfv^*)$ for all $j \neq i$. Then $h_i(\bfv^*) > 0$, and since $h_i$ attains a minimum of zero, there exists a nearby vector $\bfv_\epsilon$ such that $h_i(\bfv_\epsilon) < h_i(\bfv^*)$ and $h_j(\bfv_\epsilon) \leq h_i(\bfv_\epsilon)$ for all $j$. This contradicts optimality of $\bfv^*$. The second part follows since $z^*$ is the maximum loss.
\end{proof}

\subsection{Proof of Theorem \ref{theorem:twogroup_duality}}

\begin{proof}
    Since we are in the case $|\groups|=2$, we can consider a simplified formulation. We notice that $\mu_2=1-\mu_1$ and set $\mu_1=\mu$ and $\mu_2=1-\mu$. We also set $\bfA^1=\bfA$, $\bfA^2=\bfB$ and $\matrixC(\mu)=\mu \bfA^\top\bfA + (1-\mu) \bfB^\top\bfB$. Thus, Problem $\ref{prob:dual}$ becomes:
\begin{equation}
  \max_{\mu \in \reals} \mu s_1 + (1-\mu) s_2 - \lambda_{max}(\matrixC(\mu)), \quad \mu \in [0,1].
\end{equation}

We can now perform the standard KKT analysis. 
The dual lagrangian is: 
\begin{equation*}
\lagrangian_D (\mu, \xi_1, \xi_2) = g(\mu) + \xi_1\mu + \xi_2(1-\mu).
\end{equation*}
The stationarity condition is:
\[
\frac{\partial}{\partial \mu} \lagrangian_D (\mu^*, \xi_1, \xi_2)= \frac{\partial}{\partial \mu} g(\mu^*) + \xi_1 - \xi_2 =0. 
\]
Additionally, the complementary slackness condition requires that $\xi_1\mu=0$ and $\xi_2(1-\mu)=0$. To see this, first recall the duality between \faireig and \dual, from which we know that $\mu_1=\mu$ and $\mu_2=1-\mu$ are the associated multipliers with constraints $h_A-z$ and $h_B-z$ of \faireig. From Theorem~\ref{theorem:equal_loss} we know that $h_A-z=0$ and $h_B-z=0$ and thus from complementary slackness we can infer that $\mu$ can be neither $0$ or $1$. Similarly, complementary slackness between $\mu$ and $\xi_1$ and $\xi_2$ indicates that $\xi_1=\xi_2=0$. 

Thus, stationarity simply reduces to $\frac{\partial}{\partial \mu}g(\mu^*)=0$. From this and using equation \ref{eq:final_gradient}, it follows that: 
\begin{equation}
\label{eq:equal_mustar}
s_1-s_2-\bfv^\top(\mu^*)(\bfA^\top\bfA - \bfB^\top\bfB)\bfv(\mu^*)= 0.  
\end{equation}

Therefore, $\bfv(\mu^*)$ leads to equal loss between the two groups. 
Additionally, this stationary point is a global maximum of $g$. 
To see this, we take the second derivative of $g$:
\[
\frac{\partial^2g}{\partial\mu^2} = -\frac{\partial^2}{\partial\mu^2} \lambda_{max} (\matrixC(\mu)). 
\]

The Hadamard second variation formula \cite{schiffer1946hadamard}, gives us an analytical expression for the second derivative of $\lambda_{max}$:
\begin{align}
 &\frac{\partial^2}{\partial\mu^2} \lambda_{max}(\matrixC(\mu))=\nonumber\\
 &\parv^\top\frac{\partial^2\matrixC(\mu)}{\partial\mu^2}\parv+2\sum_{j \neq max}\frac{|\parv^\top\frac{\partial\matrixC(\mu)}{\partial\mu}\mathbf{v}_j(\mu)|}{\lambda_{max} - \lambda_j(\mu)}. \label{eq:had_second}  
\end{align}
where $\lambda_j,\mathbf{v}_j$ are eigenvalue-eigenvector pairs corresponding to smaller eigenvalues. The first term of Equation~\ref{eq:had_second} vanishes ($\matrixC(\mu)$ is only linearly dependent on $\mu$), while the numerator and denominator in the second term are trivially positive (since $\matrixC(\mu)$ is positive semidefinite and $\lambda_{max} > \lambda_j$. An important thing to note is that we have assumed simple spectrum. 
From this we can conclude that $\frac{\partial^2 g}{\partial \mu^2}<0$, i.e., the function is concave, and thus has a unique maximum, at $\mu^*$.
At $\mu^*$, we have that: 
\begin{align*}
g(\mu^*)&=s_1-\bfv(\bfmu^*)^\top\bfA^\top\bfA\bfv(\bfmu^*) \\
&=s_2-\bfv(\bfmu^*)^\top\bfB^\top\bfB\bfv(\bfmu^*).
\end{align*}
As $\bfv(\bfmu^*)$ is also a feasible point of Problem \ref{prob:faireig_optimization}, with some value $\overline{z}$, we have that $g(\mu^*)=\overline{z}$ and since the primal is always lower bounded by the dual, we conclude that strong duality holds.
\end{proof}

\begin{lemma}
\label{lemma:root_finding}
Define $q(\mu)=s_1-s_2-\bfv^\top(\mu)(\bfA^\top\bfA - \bfB^\top\bfB)\bfv(\mu)$. Then, $\mu^*$ is a root of $q(\mu)$ and additionally $q(\mu)$ is monotone with respect to $\mu$ 
\end{lemma}

The fact that $\mu^*$ is a root of $q(\mu)$ follows directly from Equation \ref{eq:equal_mustar}. The monotonicity follows from $\frac{\partial q}{\partial \mu}=-\frac{\partial^2}{\partial\mu^2} \lambda_{max} (\matrixC(\mu)) > 0$. 
This has an interesting consequence for the problem under investigation when $|\groups|=2$. The fact that a unique root exists in $\mu \in (0,1)$ and the monotonicity mean that we can resort to a root-finding algorithm (such as Brent's method~\cite{brent1971algorithm} or the bisection method~\cite{ehiwario2014comparative})  to locate the optimal $\mu^*$. In fact, as we show in the experiments, such an algorithm is highly effective for \faireig, when $|\groups|=2$. 
By default, we use the aforementioned Brent's method for finding the unique root  $\mu \in (0,1)$. 

Note that a similar approach based on root-finding algorithms 
cannot be applied to the case of more than two groups and there is no obvious way to extend this approach to the general case.

\subsection{Proof of Lemma \ref{lemma:SDP_tight} }

\begin{proof}
We begin by reformulating the dual problem from Problem~\ref{prob:dual} using the Schur complement. Recall that the dual objective is:
\[
\max_{\bfmu \in \reals^k} \quad \bfmu^\top \bfs - \lambda_{\max}(\bfA(\bfmu)),
\]
subject to $\bfone^\top \bfmu = 1$ and $\bfmu \geq 0$, where $\bfA(\bfmu) = \sum_{i=1}^k \mu_i \Ait$.

To express the $\lambda_{\max}$ constraint as a semidefinite constraint, we introduce an auxiliary scalar variable $\lambda$ such that $\lambda \geq \lambda_{\max}(\bfA(\bfmu))$. This constraint is equivalent to requiring:
\[
\bfA(\bfmu) \preceq \lambda \matrixI \quad \Longleftrightarrow \quad -\bfA(\bfmu) + \lambda \matrixI \succeq 0.
\]

We now introduce another scalar variable $\gamma$ to represent the full objective:
\[
\gamma \leq \bfmu^\top \bfs - \lambda.
\]
This can also be encoded via a semidefinite constraint using the Schur complement, leading to the block matrix:
\[
\begin{bmatrix}
-\bfA(\bfmu) + \lambda \matrixI & 0 \\
0 & \bfmu^\top \bfs - \gamma
\end{bmatrix} \succeq 0.
\]

Putting this together, the dual can now be written as the following semidefinite program (SDP):
\begin{align*}
\max_{\bfmu \in \reals^k,\, \lambda,\, \gamma} \quad & \gamma \\
\text{s.t.} \quad &
\begin{bmatrix}
-\bfA(\bfmu) + \lambda \matrixI & 0 \\
0 & \bfmu^\top \bfs - \gamma
\end{bmatrix} \succeq 0, \\
& \bfone^\top \bfmu = 1, \quad \bfmu \geq 0.
\end{align*}

We now observe that the SDP relaxation defined in Algorithm~\ref{alg:sdp} corresponds precisely to the dual of this reformulated problem, with dual variable $\matrixX$ representing a lifted version of the rank-one solution $\bfv\bfv^\top$. From our earlier analysis (and standard results in convex optimization), we know that strong duality holds between this pair of SDPs.

Hence, the semidefinite relaxation in Algorithm~\ref{alg:sdp} is tight when the dual optimum is attained and matches the primal optimum. This implies that Algorithm~\ref{alg:sdp} solves the original \faireig problem (Problem~\ref{prob:faireig_optimization}) to global optimality.
\end{proof}

\subsection{Proof of Lemma \ref{lemma:alg_equal}}

\begin{proof}
Observe that $\bfV=\{\bfv_1,\ldots,\bfv_d\}$ is a matrix with orthonormal columns since it is constructed using Algorithm \ref{alg:fairsvd}. Hence, we can invoke Property~\ref{property:projection} along with Theorem~\ref{theorem:equal_loss} to obtain the result.
Namely, after running Algorithm \ref{alg:fairsvd}, we obtain $\bfV=\{\bfv_1, \ldots, \bfv_d\}$, which gives a total error of $\sum_{i=1}^{d} \loss(\bfA,\bfv_i)$ for group $A$ and a total error of $\sum_{i=1}^{d} \loss(\bfB,\bfv_i)$ for group $B$. We know that $\loss(\bfA,\bfv_i)=\loss(\bfB,\bfv_i)$ for any $i \in \{1,\ldots,d\}$  due to Theorem \ref{theorem:equal_loss}. The lemma then follows. 

\end{proof}

\clearpage

\bibliographystyle{ACM-Reference-Format}
\bibliography{citations}

\end{document}